\documentclass[sn-mathphys]{sn-jnl}
\usepackage[utf8]{inputenc}
\usepackage{lineno}
\usepackage{hyperref}
\usepackage{booktabs}
\usepackage{xcolor}
\usepackage{amsmath}
\usepackage{algpseudocode}
\usepackage{algorithm}
\usepackage{multirow}

\usepackage{graphicx}
\usepackage[graphicx]{realboxes}
\usepackage{lscape}
\usepackage{float}
\newfloat{algorithm}{t}{lop}

\jyear{2022}
\begin{document}

% llncs title begin

% \title{Combining Monte Carlo Tree Search and Heuristic Search for Weighted Vertex Coloring}
% \titlerunning{MCTS for WVCP}

% % Cyril Grelier \orcid{0000-0002-6234-8278}
% \author{Cyril Grelier \and Olivier Goudet \and Jin-Kao Hao\thanks{Corresponding author}}
% \institute{LERIA, Université d’Angers, 2 Boulevard Lavoisier, 49045 Angers, France \email{\{cyril.grelier,olivier.goudet,jin-kao.hao\}@univ-angers.fr}}
% \authorrunning{Grelier, Goudet and Hao}

% \maketitle

% \begin{abstract}
%     This work investigates the Monte Carlo Tree Search (MCTS) method combined with dedicated heuristics for solving the Weighted Vertex Coloring Problem.
%     In addition to the basic MCTS algorithm, we study several MCTS variants where the conventional random simulation is replaced by other simulation strategies including greedy and local search heuristics.
%     We conduct experiments on well-known benchmark instances to assess these combined MCTS variants.
%     We provide empirical evidence to shed light on the advantages and limits of each simulation strategy.
%     This is an extension of the work \cite{GrelierGH22} presented at EvoCOP2022.

%     \keywords{Monte Carlo Tree Search \and local search \and graph coloring \and weighted vertex coloring.}
% \end{abstract}

% llncs title end

% sn-jnl title begin
\title[MCTS for WVCP]{Combining Monte Carlo Tree Search and Heuristic Search for Weighted Vertex Coloring}

% Cyril Grelier \orcid{0000-0002-6234-8278}
\author[1]{\fnm{Cyril} \sur{Grelier}}\email{cyril.grelier@univ-angers.fr}

\author[1]{\fnm{Olivier} \sur{Goudet}}\email{olivier.goudet@univ-angers.fr}

\author*[1]{\fnm{Jin-Kao} \sur{Hao}}\email{jin-kao.hao@univ-angers.fr}

\affil[1]{\orgdiv{LERIA}, \orgname{Université d’Angers}, \orgaddress{\street{2 Boulevard Lavoisier}, \city{Angers}, \postcode{49045}, \country{France}}}

\abstract{
    This work investigates the Monte Carlo Tree Search (MCTS) method combined with dedicated heuristics for solving the Weighted Vertex Coloring Problem.
    In addition to the basic MCTS algorithm, we study several MCTS variants where the conventional random simulation is replaced by other simulation strategies including greedy and local search heuristics.
    We conduct experiments on well-known benchmark instances to assess these combined MCTS variants.
    We provide empirical evidence to shed light on the advantages and limits of each simulation strategy.
    This is an extension of the work \cite{GrelierGH22} presented at EvoCOP2022.  The last version of this  article is published in the journal SN Computer Science (Springer). 
}


% sn-jnl title end

\maketitle

%%%%%%%%%%%%%%%%%%%%%%%%%%%%%%%%%%%%%%%%%%%%%%%%%%%%%%%%%%%%%%%%%%%%%%%%%%%%%%%%%%%%%%%%%%%%%%%%%%%%
%                                            Introduction
%%%%%%%%%%%%%%%%%%%%%%%%%%%%%%%%%%%%%%%%%%%%%%%%%%%%%%%%%%%%%%%%%%%%%%%%%%%%%%%%%%%%%%%%%%%%%%%%%%%%

\section{Introduction}

The well-known Graph Coloring Problem (GCP) is to color the vertices of a graph using as few colors as possible such that no adjacent vertices share the same color (\textit{legal} or \textit{feasible} solution).
The GCP can also be considered as partitioning the vertex set of the graph into a minimum number of color groups such that no vertices in each color group are adjacent.
The GCP has numerous practical applications in various domains \cite{lewis_guide_2021} and has been studied for a long time.

A variant of the GCP called the Weighted Vertex Coloring Problem (WVCP) has recently attracted much interest in the literature \cite{Goudet_Grelier_Hao_2021,Nogueira_Tavares_Maciel_2021,Sun_Hao_Lai_Wu_2018,Wang_Cai_Pan_Li_Yin_2020}.
In this problem, each vertex of the graph has a weight and the objective is to find a \textit{legal} solution such that the sum of the weights of the heaviest vertex of each color group is minimized.
Formally, given a weighted graph $G=(V,E)$ with vertex set $V$ ($n =$ \textbar $V$\textbar) and edge set $E$, and let $W$ be the set of weights $w(v)$ associated to each vertex $v$ in $V$, the WVCP consists in finding a partition of the vertices in $V$ into $k$ color groups $S=\{V_1,\dots,V_k\}$ ($1 \leq k \leq n)$ such that no adjacent vertices belong to the same color group and such that the score $\sum_{i=1}^{k}\max_{v\in V_i}{w(v)}$ is minimized.
One can notice that when all the weights $w(v)$ ($v \in V$) are equal to one, finding an optimal solution of this problem with a minimum score corresponds to solving the GCP.
Therefore, the WVCP can be seen as a more general problem than the GCP and is NP-hard.

\begin{figure}[!ht]
    \centering
    \includegraphics[width=0.9\textwidth,keepaspectratio]{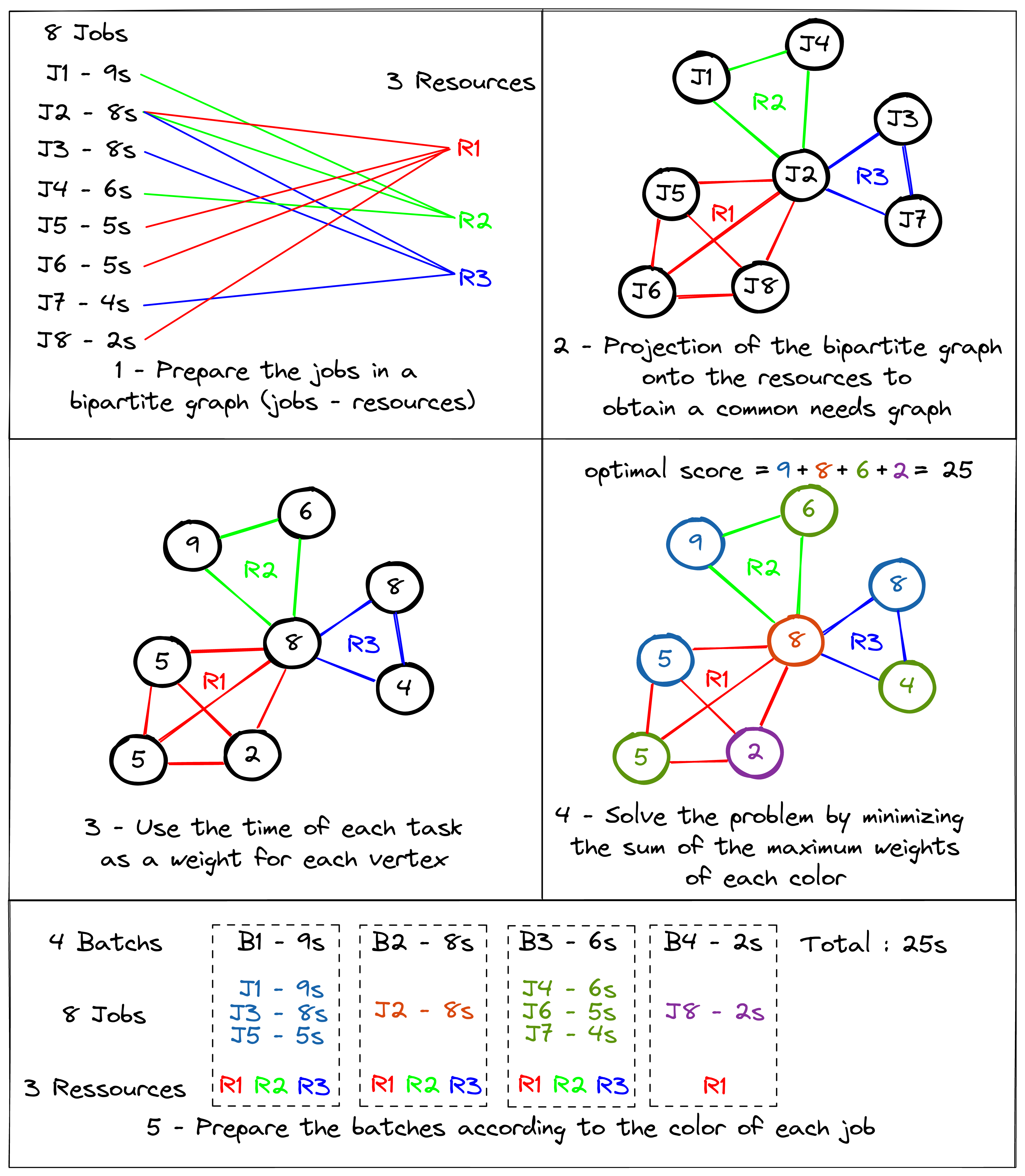}
    \caption{This figure shows an application of the WVCP for scheduling jobs into batches in a multiprocessor environment with restricted access to certain resources.}
    \label{fig:task}
\end{figure}

The WVCP is a relevant model for several applications such as matrix decomposition \cite{Prais_Ribeiro_2000}, buffer size management, and scheduling of jobs into batches in a multiprocessor environment \cite{Pemmaraju_Raman_2005}.
Let us consider the last application as illustrated in Figure \ref{fig:task}.
The objective of this scheduling problem is to execute a set of jobs in a minimum total amount of time.
There is no constraint on the number of jobs that can be run in parallel in this environment.
However, each job requires a specific execution time and exclusive access to certain resources.
Therefore, the time required to complete a batch of jobs in parallel is the time required to complete the longest job in that batch, and two jobs requiring the same resource cannot be launched in the same batch.
Solving this problem within the WVCP modeling framework can be done in five steps as displayed in Figure \ref{fig:task}: (i) a bipartite graph is used to represent the jobs and the resources required for each job; (ii) this bipartite graph is projected onto the resources to obtain a weighted graph where each vertex is a job and two jobs requiring the same resources are linked by an edge; (iii) a weight corresponding to the time needed to complete a job is set on the corresponding vertex of this graph; (iv) after solving the WVCP associated to this graph, a legal solution is found with an optimal score of 25, corresponding to the sum of the weights of the heaviest vertex of each color group; (v) this partition of vertices allows to set up a job schedule in four batches, which respects the resource constraints, and whose minimum total execution time is 25 seconds.

Different methods have been proposed in the literature to solve the WVCP.
First, this problem has been tackled with exact methods: a branch-and-price algorithm \cite{furini2012exact} and two ILP models proposed in \cite{malaguti2009models} and \cite{Cornaz_Furini_Malaguti_2017} with a transformation of the WVCP into a maximum weight independent set problem.
These exact methods can prove the optimality on small instances but fail on large instances.

To handle large graphs, several heuristics have been introduced to solve the problem approximately \cite{Nogueira_Tavares_Maciel_2021,Prais_Ribeiro_2000,Sun_Hao_Lai_Wu_2018,Wang_Cai_Pan_Li_Yin_2020}.
The first category of heuristics is based on the local search framework, which iteratively makes transitions from the current solution to a neighbor solution.
Three different approaches have been considered to explore the search space: legal, partial legal, or penalty strategies.
The legal strategy starts from a \textit{legal} solution and minimizes the score by performing only legal moves so that no color conflict is created in the new solution \cite{Prais_Ribeiro_2000}.
The partial legal strategy allows only legal coloring and keeps a set of uncolored vertices to avoid conflicts \cite{Nogueira_Tavares_Maciel_2021}.
The penalty strategy considers both legal and illegal solutions in the search space \cite{Sun_Hao_Lai_Wu_2018,Wang_Cai_Pan_Li_Yin_2020}, and uses a weighted evaluation function to minimize both the WVCP objective function and the number of conflicts in the illegal solutions.
To escape local optima traps, these local search algorithms incorporate different mechanisms such as perturbation strategies \cite{Prais_Ribeiro_2000,Sun_Hao_Lai_Wu_2018}, tabu list \cite{Nogueira_Tavares_Maciel_2021,Sun_Hao_Lai_Wu_2018} and constraint reweighting schemes \cite{Wang_Cai_Pan_Li_Yin_2020}.

The second category of existing heuristics for the WVCP relies on the population-based memetic framework that combines local search with crossovers.
A recent algorithm \cite{Goudet_Grelier_Hao_2021} of this category uses a deep neural network to learn an invariant by color permutation regression model, useful to select the most promising crossovers at each generation.

Research on combining such learning techniques and heuristics has received increasing attention in the past years \cite{zhou2018improving,zhou2020frequent}.
In these new frameworks, useful information (e.g., relevant patterns) is learned from past search trajectories and used to guide a local search algorithm.

This study continues on this path and investigates the potential benefits of combining Monte Carlo Tree Search (MCTS) and sequential coloring or local search algorithms for solving the WVCP.
MCTS is a heuristic search algorithm that generated considerable interest due to its spectacular success for the game of Go \cite{gelly2006modification}, and in other domains (see the survey \cite{browne2012survey} on this topic).
It has been recently revisited in combination with modern deep learning techniques for difficult two-player games (cf. AlphaGo \cite{Silver_Huang_Maddison_2016}).
MCTS has also been applied to combinatorial optimization problems seen as a one-player game such as the traveling salesman problem \cite{edelkamp2014solving} or the knapsack problem \cite{Jooken_Leyman_2020}.
An algorithm based on MCTS has recently been implemented with some success for the GCP in \cite{Cazenave_Negrevergne_Sikora_2020}.
In this work, we investigate for the first time the MCTS approach for solving the WVCP.

In MCTS, a tree is built incrementally and asymmetrically.
For each iteration, a tree policy balancing exploration and exploitation is used to find the most critical node to expand.
A simulation is then run from the expanded node and the search tree is updated with the result of this simulation.
Its incremental and asymmetric properties make MCTS a promising candidate for the WVCP because in this problem only the heaviest vertex of each color group has an impact on the objective score.
Therefore learning to color the heaviest vertices of the graph before coloring the rest of the graph seems particularly relevant for this problem.
The contributions of this work are summarized as follows.

First, we present a MCTS algorithm dedicated to the WVCP, which considers the problem from the perspective of sequential coloring with a predefined vertex order.
The exploration of the tree is accelerated with the use of specific pruning rules, which offer the possibility to explore the whole tree in a reasonable amount of time for small instances and to obtain optimality proofs.
Secondly, for large instances, when obtaining an exact result is impossible in a reasonable time, we study how this MCTS algorithm can be tightly coupled with other heuristics.
Specifically, we investigate the integration of different greedy coloring strategies and local search procedures within the MCTS algorithm.

The rest of the paper is organized as follows.
Section \ref{sec:constructionTree} introduces the weighted vertex coloring problem and the constructive approach with a tree.
Section \ref{sec:mcts_classic} describes the MCTS algorithm devised to tackle the problem.
Section \ref{sec:mcts_improved} presents the coupling of MCTS with local search.
Section \ref{sec:experimentation} reports computational results of different versions of MCTS.
Section \ref{sec:conclusion} discusses the contributions and presents research perspectives.

%%%%%%%%%%%%%%%%%%%%%%%%%%%%%%%%%%%%%%%%%%%%%%%%%%%%%%%%%%%%%%%%%%%%%%%%%%%%%%%%%%%%%%%%%%%%%%%%%%%%
%                                            Tree construction
%%%%%%%%%%%%%%%%%%%%%%%%%%%%%%%%%%%%%%%%%%%%%%%%%%%%%%%%%%%%%%%%%%%%%%%%%%%%%%%%%%%%%%%%%%%%%%%%%%%%

\section{Constructive approach with a tree for the weighted graph coloring problem \label{sec:constructionTree}}

This section presents a tree-based approach for the WVCP, which aims to explore the partial and legal search space of this problem.

\subsection{Partial and legal search space \label{sec:search_space}}

The search space $\Omega$ studied in our algorithm concerns legal, but potentially partial, $k$-colorings.
A partial legal $k$-coloring $S$ is a partition of the set of vertices $V$ into $k$ disjoint independent sets $V_i$ ($1 \leq i \leq k$), and a set of uncolored vertices $U = V \backslash \bigcup_{i=1}^{k}V_i$.
A independent set $V_i$ is a set of mutually non adjacent vertices of the graph: $\forall u,v \in V_i, (u,v) \notin E$.
For the WVCP, the number of colors $k$ that can be used is not known in advance.
Nevertheless, it is not lower than the chromatic number of the graph $\chi(G)$ and not greater than the number of vertices $n$ of the graph.
A solution of the WVCP is denoted as partial if $U \neq \emptyset$ and complete otherwise.
The objective of the WVCP is to find a complete solution $S$ with a minimum score $f(S)$ given by: $f(S) = \sum_{i=1}^{k}\max_{v\in V_i}{w(v)}.$

\subsection{Tree search for weighted vertex coloring} \label{sec:treeSearch}

Backtracking-based tree search is a popular approach for the graph coloring problem \cite{brelaz1979new,KubaleJ84,lewis_guide_2021}.
In our case, a tree search algorithm can be used to explore the partial and legal search space of the WVCP previously defined.

Starting from a solution where no vertex is colored (i.e., $U=V$) and that corresponds to the root node $R$ of the tree, child nodes $C$ are successively selected in the tree, consisting of coloring one new vertex at a time.
This process is repeated until a terminal node $T$ is reached (all the vertices are colored).
A complete solution (i.e., a legal coloring) corresponds thus to a branch from the root node to a terminal node.

The selection of each child node corresponds to applying a move to the current partial solution being constructed.
A move consists of assigning a particular color $i$ to an uncolored vertex $u \in U$, denoted as $<u,U,V_i>$.
Applying a move to the current partial solution $S$, results in a new solution $S\ \oplus <u,U,V_i>$.
This tree search algorithm only considers legal moves to stay in the partial legal space.
For a partial solution $S=\{V_1,...,V_k,U\}$, a move $<u,U,V_i>$ is said legal if no vertex of $V_i$ is adjacent to the vertex $u$.
At each level of the tree, there is at least one possible legal move that applies to a vertex a new color that has never been used before (or putting this vertex in a new empty set $V_{i}$, $k+1 \leq i \leq n$).

Applying a succession of $n$ legal moves from the initial solution results in a legal coloring of the WVCP and reaches a terminal node of the tree.
During this process, at the level $t$ of the tree ($0 \leq t < n)$, the current legal and partial solution $S=\{V_1,...,V_k,U\}$ has already used $k$ colors and $t$ vertices have already received a color.
Therefore \textbar$U$\textbar $= n - t$.

At this level, a first naive approach could be to consider all the possible legal moves, corresponding to choosing a vertex in the set $U$ and assigning to the vertex a color $i$, with $1 \leq i \leq n$.
This kind of choice can work with small graphs but with large graphs, the number of possible legal moves becomes huge.
Indeed, at each level $t$, the number of possible legal moves can go up to $(n - t) \times n$.
To reduce the set of move possibilities, we consider the vertices of the graph in a predefined order $(v_1,\dots,v_{n})$.
Moreover, to choose a color for the incoming vertex, we consider only the colors already used in the partial solution plus one color (creation of a new independent set).
Thus for a current legal and partial solution $S=\{V_1,...,V_k,U\}$, at most $k+1$ moves are considered.
The set of legal moves is:
\begin{equation}
    \mathcal{L}(S) = \{ <u,U,V_i>, 1 \leq i \leq k, \forall v \in V_i, (u,v) \notin E \}\ \cup <u,U,V_{k+1}> \label{eq:set_legal_moves}
\end{equation}
This decision cuts the symmetries in the tree while reducing the number of branching factors at each level of the tree.

\subsection{Predefined vertex order}\label{sec:order}

We propose to consider a predefined ordering of the vertices, sorted by weight and then by degree.
Vertices with higher weights are placed first.
If two vertices have the same weight, then the vertex with the higher degree is placed first.
This order is intuitively relevant for the WVCP because it is more important to place first the vertices with heavy weights which have the most impact on the score as well as the vertices with the highest degree because they are the most constrained decision variables.
Such ordering has already been shown to be effective with greedy constructive approaches for the GCP \cite{brelaz1979new} and the WVCP \cite{Nogueira_Tavares_Maciel_2021}.

Moreover, this vertex ordering allows a simple score calculation while building the tree.
Indeed, as the vertices are sorted by descending order of their weights, and the score of the WVCP only counts the maximum weight of each color group, with this vertex order, the score only increases by the value $w(v)$ when a new color group is created for the vertex $v$.

%%%%%%%%%%%%%%%%%%%%%%%%%%%%%%%%%%%%%%%%%%%%%%%%%%%%%%%%%%%%%%%%%%%%%%%%%%%%%%%%%%%%%%%%%%%%%%%%%%%%
%                                               MCTS
%%%%%%%%%%%%%%%%%%%%%%%%%%%%%%%%%%%%%%%%%%%%%%%%%%%%%%%%%%%%%%%%%%%%%%%%%%%%%%%%%%%%%%%%%%%%%%%%%%%%

\section{Monte Carlo Tree Search for weighted vertex coloring} \label{sec:mcts_classic}

The search tree presented in the last subsection can be huge, in particular for large instances.
Therefore, in practice, it is often impossible to perform an exhaustive search of this tree, due to expensive computing time and memory requirements.
We turn now to an adaptation of the MCTS algorithm for the WVCP to explore this search tree.
MCTS keeps in memory a tree (hereinafter referred to as the MCTS tree) that only corresponds to the already explored nodes of the search tree presented in the last subsection.
In the MCTS tree, a leaf is a node whose children have not yet all been explored while a terminal node corresponds to a complete solution.
MCTS can guide the search toward the most promising branches of the tree, by balancing exploitation and exploration and continuously learning at each iteration.

\subsection{General framework}

The MCTS algorithm for the WVCP is shown in Algorithm \ref{alg:MCTS}.
The algorithm takes a weighted graph as input and tries to find a legal coloring $S$ with the minimum score $f(S)$.
The algorithm starts with an initial solution where the first vertex is placed in the first color group.
This is the root node of the MCTS tree.
Then, the algorithm repeats several iterations until a stopping criterion is met.
At every iteration, one legal solution is completely built, which corresponds to walking along a path from the root node to a leaf node of the MCTS tree and performing a simulation (or playout/rollout) until a terminal node of the search tree is reached (when all vertices are colored).

Each iteration of the MCTS algorithm involves the execution of 5 steps to explore the search tree with legal moves (cf. Section \ref{sec:constructionTree}):

\begin{enumerate}
    \item \textbf{Selection} From the root node of the MCTS tree, successive child nodes are selected until a leaf node is reached.
          The selection process balances the exploration-exploitation trade-off.
          The exploitation score is linked to the average score obtained after having selected this child node and is used to guide the algorithm to a part of the tree where the scores are the lowest (the WVCP is a minimization problem).
          The exploration score is linked to the number of visits to the child node and will incite the algorithm to explore new parts of the tree, which have not yet been explored.
    \item \textbf{Expansion} The MCTS tree grows by adding a new child node to the leaf node reached during the selection phase.
    \item \textbf{Simulation} From the newly added node, the current partial solution is completed with legal moves, randomly or by using heuristics.
    \item \textbf{Update} After the simulation, the average score and the number of visits of each node on the explored branch are updated.
    \item \textbf{Pruning} If a new best score is found, some branches of the MCTS tree may be pruned if it is not possible to improve the best current score with it.
\end{enumerate}

\noindent The algorithm continues until one of the following conditions is reached:
\begin{itemize}
    \item there are no more child nodes to expand, meaning the search tree has been fully explored.
          In this case, the best score found is proven to be optimal.
    \item a cutoff time is attained.
          The minimum score found so far is returned.
          It corresponds to an upper bound of the score for the given instance.
\end{itemize}

\begin{algorithm} \caption{MCTS algorithm for the WVCP\label{alg:MCTS}}
    \begin{algorithmic}[1]
        \State \bf{Input}: \normalfont{Weighted graph $G = (V, W, E)$}
        \State \bf{Output}: \normalfont{The best legal coloring $S^*$ found}
        \State $S^*=\emptyset$ and $f(S^*) = MaxInt$
        \While{stop condition is not met}
        \State $C \leftarrow R$ \Comment{Current node corresponding to the root node of the tree}
        \State $S$ $\leftarrow$ $\{V_1,U\}$ with $V_1 = \{v_1\}$ and $U = V \backslash V_1$ \Comment{Current solution initialized with the first vertex in the first color group}
        \State
        /* Selection */ \Comment{Section \ref{mcts:selection}}
        \While{$C$ is not a leaf}
        \State $C$ $\leftarrow$ select\_best\_child(C) with legal move $<u,U,V_i>$
        \State $S \leftarrow S\ \oplus <u,U,V_i>$
        \EndWhile
        \State
        /* Expansion */ \Comment{Section \ref{mcts:expansion}}
        \If{C has a potential child, not yet open}
        \State $C$ $\leftarrow$ open\_first\_child\_not\_open(C) with legal move $<u,U,V_i>$
        \State $S \leftarrow S\ \oplus <u,U,V_i>$
        \EndIf
        \State
        /* Simulation */ \Comment{Section \ref{mcts:simulation}}
        \State complete\_partial\_solution($S$)
        \State
        /* Update */ \Comment{Section \ref{mcts:update}}
        \While{$C \neq R$}
        \State update(C,f(S))
        \State $C \leftarrow$ parent(C)
        \EndWhile
        \If{$f(S) < f(S^*)$}
        \State $S^* \leftarrow S$
        \State /* Pruning */ \Comment{Section \ref{mcts:pruning}}
        \State apply pruning rules
        \EndIf
        \EndWhile
        \State return $S^*$
    \end{algorithmic}
\end{algorithm}

\subsection{Selection} \label{mcts:selection}

The selection starts from the root node of the MCTS tree and selects children nodes until a leaf node is reached.
At every level $t$ of the MCTS tree, if the current node $C_t$ is corresponds to a partial solution $S=\{V_1,...,V_k,U\}$ with $t$ vertices already colored and $k$ colors used, there are $l$ possible legal moves, with $1 \leq l \leq k+1$.
Therefore, from the node $C_t$, $l$ potential children $C^1_{t+1}, \dots, C^l_{t+1}$ can be selected.

If $l > 1$, the selection of the most promising child node can be seen as a multi-armed bandit problem \cite{Lai_Robbins_1985} with $l$ levers.
This problem of choosing the next node can be solved with the UCT algorithm for Monte Carlo tree search by selecting the child with the maximum value of the following expression \cite{Jooken_Leyman_2020}:
\begin{equation}
    normalized\_score(C^i_{t+1}) + c \times \sqrt{\frac{2* ln(nb\_visits(C_t))}{nb\_visits(C^i_{t+1})}},\ \text{for}\ 1 \leq i \leq l.
    \label{eq:score_UCT}
\end{equation}
\noindent Here, $nb\_visits(C)$ corresponds to the number of times the node $C$ has been chosen to build a solution.
$c$ is a real positive coefficient allowing to balance the compromise between exploitation and exploration.
It is set by default to the value of one\footnote{A sensitivity analysis of this important hyperparameter is shown in Section \ref{sec:coeff_c}.}.
$normalized\_score(C^i_{t+1})$ corresponds to a normalized score of the child node $C^i_{t+1}$ ($1 \leq i \leq l$) given by:
$$normalized\_score(C^i_{t+1})= \frac{rank(C^i_{t+1})}{\sum_{i=1}^l rank(C^i_{t+1})},$$

\noindent where $rank(C^i_{t+1})$ is defined as the rank between 1 and $l$ of the nodes $C^i_{t+1}$ obtained by sorting from bad to good according their average values $avg\_score(C^i_{t+1})$ (nodes that seem more promising get a higher score).
$avg\_score(C^i_{t+1})$ is the mean score on the sub-branch with the node $C^i_{t+1}$ selected obtained after all previous simulations.

\subsection{Expansion} \label{mcts:expansion}

From the node $C$ of the MCTS tree reached during the selection procedure, one new child of $C$ is open and its corresponding legal move is applied to the current solution.
Among the unopened children, the node associated with the lowest color number $i$ is selected.
Therefore the child node needing the creation of a new color (and increasing the score) will be selected last.

\subsection{Simulation} \label{mcts:simulation}

The simulation takes the current partial and legal solution found after the expansion phase and colors the remaining vertices.
In the original MCTS algorithm, the simulation consists in choosing random moves in the set of all legal moves $\mathcal{L}(S)$ defined by equation (\ref{eq:set_legal_moves}) until the solution is completed.
We call this first version \textit{MCTS+Random}.
As shown in the experimental section, this version is not very efficient as the number of colors grows rapidly.
Therefore, we propose two other simulation procedures:
\begin{itemize}
    \item a constrained greedy algorithm that chooses a legal move prioritizing the moves which do not locally increase the score of the current partial solution $S=\{V_1,...,V_k,U\}$:
          \begin{equation}
              \mathcal{L}^g(S) = \{ <u,U,V_i>, 1 \leq i \leq k, \forall v \in V_i, (u,v) \notin E \}
              \label{eq:set_legal_moves_no_new_color}
          \end{equation}
          It only chooses the move $<u,U,V_{k+1}>$, consisting in opening a new color group and increasing the current score by $w(u)$, only if $\mathcal{L}^g(S) = \emptyset$.
          We call this version \textit{MCTS+Greedy-Random}.
    \item a greedy deterministic procedure which always chooses a legal move in $\mathcal{L}(S)$ with the first available color $i$.
          We call this version \textit{MCTS+Greedy}.
\end{itemize}

\subsection{Update} \label{mcts:update}

Once the simulation is over, a complete solution of the WVCP is obtained.
If this solution is better than the best recorded solution found so far $S^*$ (i.e., $f(S) < f(S^*)$), $S$ becomes the new global best solution $S^*$.

Then, a backpropagation procedure updates each node $C$ of the whole branch of the MCTS tree which has led to this solution:
\begin{itemize}
    \item the running average score of each node $C$ of the branch is updated with the score $f(S)$:
          \begin{equation}
              avg\_score(C) \leftarrow \frac{avg\_score(C) \times nb\_visits(C) + f(S)}{nb\_visits(C) + 1}
          \end{equation}
    \item the counter of visits $nb\_visits(C)$ of each node of the branch is increased by one.
\end{itemize}

\subsection{Pruning} \label{mcts:pruning}

During an iteration of MCTS, three pruning rules are applied:

\begin{enumerate}
    \item during expansion: if the score $f(S)$ of the partial solution associated with a node visited during this iteration of MCTS is equal or higher to the current best-found score $f(S^*)$, then the node is deleted as the score of a partial solution cannot decrease when more vertices are colored.
    \item when the best score $f(S^*)$ is found, the tree is \textit{cleaned}.
          A heuristic goes through the whole tree and deletes children and possible children associated with a partial score $f(S)$ equal or superior to the best score $f(S^*)$.
    \item if a node is \textit{completely explored}, it is deleted and will not be explored in the MCTS tree anymore.
          A node is said \textit{completely explored} if it is a leaf node without children, or if all of its children have already been opened once and have all been deleted.
          Note that this third pruning step is recursive as a node deletion can result in the deletion of its parent if it has no more children, and so on.
\end{enumerate}

These three pruning rules and the fact that the symmetries are cut in the tree by restricting the set of legal moves considered at each step (see Section \ref{sec:treeSearch}) offer the possibility to explore the whole tree in a reasonable amount of time for small instances.
This peculiarity of the algorithm makes it possible to obtain an optimality proof for such instances.

\subsection{Toy example}

\begin{figure}[!hbt]
    \centering
    \includegraphics[width=0.8\textwidth,keepaspectratio]{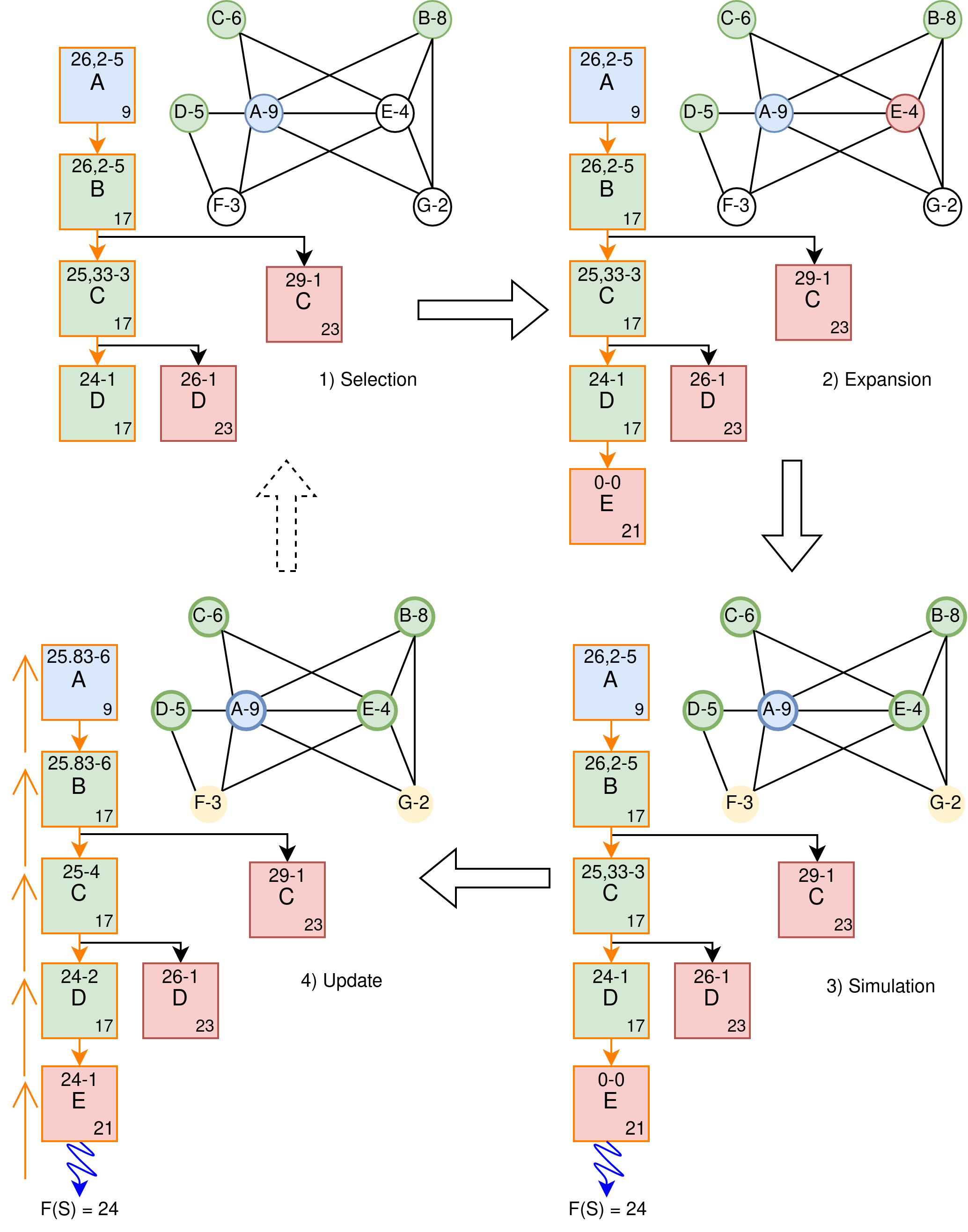}
    \caption{Toy example of one MCTS iteration}
    \label{img:start}
\end{figure}

Figure \ref{img:start} displays one iteration of MCTS for the WVCP on a small graph with seven vertices named A--G with different weights between 2 and 9.
On each diagram is displayed the current state of the partial coloring solution being constructed (right) and the current state of the search tree (left).
In the search tree, each square represents a node and the number on the bottom right of a square is the score of the corresponding partial solution.
On top of each square are written the average score and the number of visits of each node.
In addition to the root node (vertex A colored in blue), five nodes have already been opened in the search tree (five iterations of MCTS).
The sixth iteration of MCTS proceeds as follows.

\begin{figure}[!hbt]
    \centering
    \includegraphics[width=0.6\textwidth,keepaspectratio]{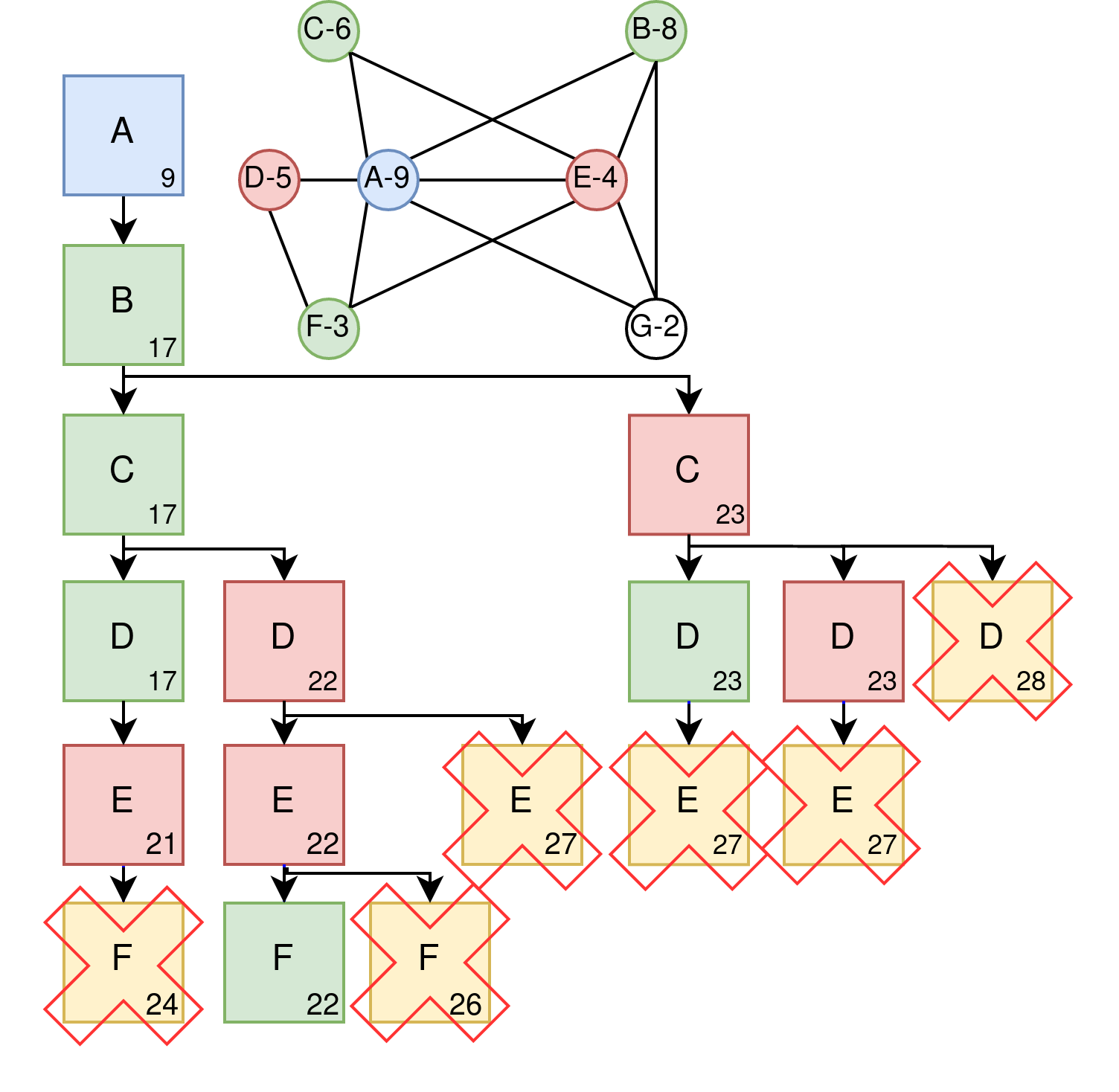}
    \caption{Toy example of the search tree pruning}
    \label{img:prune}
\end{figure}

\begin{itemize}
    \item \textbf{Selection}: from the root node, the only possible child corresponding to the vertex B in green is selected.
          From there, there are two options as vertex C can be colored in green or red.
          The most interesting option is chosen (vertex C in green) regarding the score and the number of visits of each child (cf. equation (\ref{eq:score_UCT})).
          Then, the most promising leaf is selected (D in green).
    \item \textbf{Expansion}: From the node D in green, a new node is added to the tree.
          It corresponds to E in red (as it cannot take the color blue nor green).
    \item \textbf{Simulation}: From there, the solution is completed with a greedy algorithm to obtain a complete legal solution with a score of 24.
    \item \textbf{Update}: this score of 24 is back-propagated on the explored branch (update of the average score and the number of visits of each node in the branch).
    \item \textbf{Pruning}: Figure \ref{img:prune} presents the state of the tree after some iterations.
          As the best-found score is 24, every branch of the tree with a score upper or equal to 24 is deleted (indicated with a red cross).
\end{itemize}

%%%%%%%%%%%%%%%%%%%%%%%%%%%%%%%%%%%%%%%%%%%%%%%%%%%%%%%%%%%%%%%%%%%%%%%%%%%%%%%%%%%%%%%%%%%%%%%%%%%%
%                                            MCTS + LS
%%%%%%%%%%%%%%%%%%%%%%%%%%%%%%%%%%%%%%%%%%%%%%%%%%%%%%%%%%%%%%%%%%%%%%%%%%%%%%%%%%%%%%%%%%%%%%%%%%%%

\section{Combining MCTS with Local Search} \label{sec:mcts_improved}

We now explore the possibility of improving the MCTS algorithm with local search.
Coupling MCTS with a local search algorithm is motivated by the fact that after the simulation phase, the complete solution obtained can be close to a still better solution in the search space that could be discovered by local search.
In this work, we present the coupling of MCTS with a baseline tabu search (TW) created for this work, as well as three \textit{state of the art} local search algorithms, dedicated for the WVCP: AFISA \cite{Sun_Hao_Lai_Wu_2018}, RedLS \cite{Wang_Cai_Pan_Li_Yin_2020} and ILSTS \cite{Nogueira_Tavares_Maciel_2021}.

During the simulation phase, the solution is first completed with a greedy algorithm and then improved by the local search procedure.
Note that in the first version of this work published in \cite{GrelierGH22}, to stay consistent with the search tree learned by MCTS, we allowed the local search procedure to only move the vertices of the complete solution $S$ which are still uncolored after the selection and expansion phases.
However, we have realized in the meantime that blocking vertices for the local search can lead to a lot of time spent checking for blocked vertices in the complex neighborhood explored by the various local search procedures.
It also leads to missing good opportunities to move in the search space.
Therefore, in the new version of the algorithm presented in this paper, a more efficient version of the algorithm is presented where the vertices are not frozen during the local search.
In this new version, when coupling MCTS with a local search algorithm, the resulting heuristic can be seen as an algorithm that attempts to learn a good starting point for the local search procedure, by selecting different best promising backbones of partial solutions in every iteration during the selection phase.

As one iteration does not have the same meaning for each local search, we use a time limit of $t = 0.02 \times n$ seconds to perform the search, depending on the number of vertices $n$ in the given instance.
Once the local search procedure has reached the time limit, the score corresponding to the best legal solution obtained by the local search procedure is used to update all the nodes of the branch which has led to the simulation initiation.
In the following subsections, the four different local search procedures used in this work are presented.

\subsection{Basic tabu search}

The first local search algorithm tested is a simple tabu search, named TabuWeight (TW), inspired by the classical TabuCol algorithm for the GCP \cite{Hertz_Werra_1987}.
Starting from a legal solution, TW improves it iteratively by using a \textit{one move} operator, which consists in moving a vertex from its color group to another color group, without creating conflicts.
At each iteration, the best \textit{one move} which is not forbidden by the tabu list is selected.
Each time a move is performed, the reverse move is added to the tabu list and forbidden for the next $tt$ iterations where $tt$ is a parameter called tabu tenure.
A tabu move can still be applied exceptionally if it leads to a solution, which is better than the best solution found so far (aspiration criterion).

\subsection{Adaptive feasible and infeasible tabu search}

AFISA \cite{Sun_Hao_Lai_Wu_2018} is a tabu search algorithm, which explores the candidate solutions by oscillating between illegal and legal search spaces\footnote{The illegal search space consists of solutions with conflicts (some adjacent vertices in the solution have the same color), while the legal search space consists of solutions without any conflicts.}.
To prevent the search from going too far from legal boundaries, AFISA uses a controlling coefficient to adaptively makes the algorithm go back and forth between illegal and legal spaces.
The controlling coefficient encourages the algorithm to handle in priority the vertices in conflicts before trying to reduce the WVCP score.
AFISA uses the popular \textit{one move} operator to explore the search space.

\subsection{Local search with multiple operators}

Like AFISA, RedLS \cite{Wang_Cai_Pan_Li_Yin_2020} explores the illegal and legal search spaces.
This algorithm uses the configuration checking strategy \cite{cai_local_2011} that applies multiple improvements and perturbation strategies to explore the search space.
At each iteration, RedLS perturbs the solution by moving all the heaviest vertices from one color group to another group, before minimizing the number of conflicts to recover a new legal solution.
It uses different variants of the \textit{one move} operator to reduce the number of conflicts while keeping the WVCP score as low as possible.
Each conflicting edge has a weight that is increased each time it is not resolved, to give priority to its resolution for the next iterations.

\subsection{Iterated local search with tabu search}

ILSTS \cite{Nogueira_Tavares_Maciel_2021} explores the legal and partial search spaces.
From a complete solution, the ILSTS algorithm iteratively performs 2 steps: (i) it deletes the heaviest vertices from 1 to 3 color groups $V_i$ and places them in the set of uncolored vertices $U$; (ii) it improves the solution (i.e., minimizes the score $f(S)$) by applying different variants of the \textit{one move} operator and the so-called \textit{grenade} operator until the set of uncolored vertices $U$ becomes empty.
The \textit{grenade} operator $grenade(u,V_i)$ consists in moving a vertex $u$ to $V_i$, but first, each adjacent vertices of $u$ in $V_i$ is relocated to other color groups or in $U$ to keep a legal solution.

%%%%%%%%%%%%%%%%%%%%%%%%%%%%%%%%%%%%%%%%%%%%%%%%%%%%%%%%%%%%%%%%%%%%%%%%%%%%%%%%%%%%%%%%%%%%%%%%%%%%
%                                            Experimentation
%%%%%%%%%%%%%%%%%%%%%%%%%%%%%%%%%%%%%%%%%%%%%%%%%%%%%%%%%%%%%%%%%%%%%%%%%%%%%%%%%%%%%%%%%%%%%%%%%%%%

\section{Experimentation} \label{sec:experimentation}

This section first describes the experimental settings used in this work.
Secondly, we experimentally verify the impacts of the different greedy coloring strategies used during the MCTS simulation phase.
Thirdly, an analysis of exploration versus exploration is performed.
Lastly, the relevance of coupling MCTS with a local search procedure is studied.

\subsection{Experimental settings and benchmark instances}

A total of 188 instances are used for the experimental studies: 30 rxx graphs and 35 pxx graphs from matrix decomposition \cite{Prais_Ribeiro_2000} and 123 from the DIMACS and COLOR competitions.
Two pre-processing procedures were applied to reduce the graphs of the different instances.
The first one comes from \cite{Wang_Cai_Pan_Li_Yin_2020}: if the weight of a vertex of degree $d$ is lower than the weight of the $d+1$th heaviest vertex from any clique of the graph, then the vertex can be deleted without changing the optimal WVCP score of this instance.
The second one comes from \cite{Cheeseman_Kanefsky_Taylor_1991} and is adapted to our problem: if all neighbors of a vertex $v1$ are all neighbors with a vertex $v2$ and the weight of $v2$ is greater or equal to the weight of $v1$, then $v1$ can be deleted as it can take the color of $v2$ without impacting the score.
All the original and reduced instances are available at \url{https://github.com/Cyril-Grelier/gc_instances}.

All presented algorithms are coded in C++, compiled, and optimized with the g++ 12.1 compiler.
The source code of our algorithm (and reproduced local searches) is available at \url{https://github.com/Cyril-Grelier/gc_wvcp_mcts} with complete spreadsheets of the results.
To solve each instance, 20 independent runs were performed on a computer equipped with an Intel Xeon ES 2630, 2,66 GHz CPU with a time limit of one hour, except for the exploration vs. exploitation coefficient tests where 5 to 15 hours were used.
Running the DIMACS Machine Benchmark procedure dfmax\footnote{\url{http://archive.dimacs.rutgers.edu/pub/dsj/clique/}} on our computer took 8.94 seconds to solve the instance r500.5 using gcc 12.1 without optimization flag.

In the following subsections, summary tables allowing general comparisons between the different versions of the algorithms are presented.
Detailed results on each specific instance are reported in appendices \ref{app:greedy} and \ref{app:ls}.

The 188 instances have been separated into four sets: (i) \textbf{pxx}, with the 35 pxx instances from \cite{Prais_Ribeiro_2000}, (ii) \textbf{rxx}, with the 30 rxx instances from \cite{Prais_Ribeiro_2000}, (iii) \textbf{DIMACS\_easy}, corresponding to the \textit{easy} 59 DIMACS and COLOL instances which were solved optimally by the exact algorithm MWSS \cite{Cornaz_Furini_Malaguti_2017} or the MCTS greedy variants, reported in \cite{Nogueira_Tavares_Maciel_2021} (launched with a time limit of 10 hours for each instance), and (iv) \textbf{DIMACS\_hard}, the 64 \textit{hard} DIMACS instances which have never been solved optimally in the literature.

For all the different versions of the MCTS algorithm, the coefficient $c$, allowing to balance the compromise between exploitation and exploration is set to the value of one (cf. equation (\ref{eq:score_UCT})).
A sensitivity analysis of this important hyperparameter is conducted in Section \ref{sec:coeff_c}.

\subsection{Monte Carlo Tree Search with greedy strategies}
\label{sec:MCTS_greedy}

Table \ref{tab:mcts_greedy} summaries the results of MCTS with greedy heuristics for its simulation (cf. Section \ref{mcts:simulation}).
Columns 1 and 2 show the instance sets and their size (a star is added if all the instances of the set have optimally been solved).
Columns 3-8 show the number of instances for which each method can achieve the Best Known Score (BKS) from the literature.
Some of the BKS come from \cite{Nogueira_Tavares_Maciel_2021} (1h runs) or \cite{Goudet_Grelier_Hao_2021} (many hours of calculation), otherwise, they come from our reproductions of the state of the art algorithms.
The numbers indicated with a star and in parenthesis in this table correspond to the number of times the method can prove that the best score obtained is optimal.

Column 3 shows the results of a random algorithm (R) which colors each vertex one by one with a random color.
Column 4 presents the results of the MCTS+Random algorithm with the random simulation (MCTS+R).
Column 5 reports the results of the greedy random procedure alone (GR).
This GR procedure colors each vertex one by one with a random already used color and opens a new color when it is mandatory.
Column 6 reports the results of the MCTS algorithm using this simulation strategy (MCTS+GR).
Column 7 corresponds to the deterministic greedy strategy (G).
This strategy consists in always choosing the first available color for each vertex and opening new colors only if needed.
Column 8 shows the results of MCTS+Greedy coupling MCTS with this deterministic greedy strategy (MCTS+G).

\begin{table}[h]
    \centering
    \caption{
        Summary of the number of times the Best Known Score is reached for each algorithm.
        The values with a star indicate the number of times a score has been proved optimal.
        Values in bold highlight the best results for each line.
    }
    \label{tab:mcts_greedy}
    \resizebox{\textwidth}{!}{
        \begin{tabular}{cc|cccccc}
            instances    & \textbar I\textbar & R & \multicolumn{1}{c|}{MCTS+R}   & GR & \multicolumn{1}{c|}{MCTS+GR}            & G  & MCTS+G            \\ \hline
            pxx          & 35*                & 1 & \multicolumn{1}{c|}{34 (25*)} & 13 & \multicolumn{1}{c|}{\textbf{35} (25*)}  & 13 & \textbf{35} (25*) \\
            rxx          & 30*                & 0 & \multicolumn{1}{c|}{1}        & 0  & \multicolumn{1}{c|}{\textbf{20}}        & 2  & 11                \\
            DIMACS\_easy & 59*                & 3 & \multicolumn{1}{c|}{32 (19*)} & 11 & \multicolumn{1}{c|}{\textbf{45} (20*)}  & 8  & 35 (19*)          \\
            DIMACS\_hard & 64                 & 0 & \multicolumn{1}{c|}{7}        & 2  & \multicolumn{1}{c|}{\textbf{8}}         & 1  & 6                 \\ \hline
            Total        & 188                & 4 & \multicolumn{1}{c|}{74 (44*)} & 26 & \multicolumn{1}{c|}{\textbf{108} (45*)} & 24 & 87 (44*)
        \end{tabular}
    }
\end{table}

First, we observe that all the MCTS variants dominate the baseline greedy algorithms (R, G, and GR) in terms of the number of the BKS obtained, highlighting the relevance of combining the MCTS framework and search heuristics.

With the coupled MCTS algorithms, almost all pxx instances are solved.
The rxx instances are more difficult to solve except for the version MCTS+Greedy-Random.
The instances from DIMACS\_easy are partially solved by each MCTS variant.
The instances from DIMACS\_hard show a real difficulty for all these MCTS variants, because very few methods reach the BKS of the literature.

To better compare these different algorithms, and not only relying on the number of best-known scores achieved (that can sometimes be found by "chance"), we performed pairwise comparisons between the algorithms based on the average scores obtained on each instance as displayed in Table \ref{tab:mcts_greedy_diff}.

In Table \ref{tab:mcts_greedy_diff}, the numbers in each row correspond to the number of instances for which the method is significantly better than another (with a maximum of 188 instances).
A method is said significantly better than another on a given instance if its average score measured over 20 runs is significantly better (t-test with a p-value below 0.001).
The column \textit{Total} corresponds to the number of times a method is better than another.

\begin{table}[h]
    \centering
    \caption{
        Comparison between all greedy and MCTS variants.
        As an example, the row for MCTS+Random means that the method is better on 188 instances compared to the random procedure (R), on 165 instances compared to the Greedy-Random procedure (GR), and never compared to MCTS+Greedy-Random (MCTS+GR).
        Values in bold highlight the highest value between two methods, the variant MCTS+Greedy is more often (49 times) significantly better than MCTS+Greedy-Random (which is better than MCTS+Greedy 25 times).
    }
    \label{tab:mcts_greedy_diff}
    \begin{tabular}{c|cc|cc|cc|c}
                & \rotatebox[origin=c]{90}{R} & \rotatebox[origin=c]{90}{MCTS+R} & \rotatebox[origin=c]{90}{GR} & \rotatebox[origin=c]{90}{MCTS+GR} & \rotatebox[origin=c]{90}{G} & \rotatebox[origin=c]{90}{MCTS+G} & Total \\ \hline
        R       & -                           & 0                                & 0                            & 0                                 & 0                           & 0                                & 0/5   \\
        MCTS+R  & \textbf{188}                & -                                & \textbf{165}                 & 0                                 & \textbf{95}                 & 1                                & 3/5   \\ \hline
        GR      & \textbf{188}                & 11                               & -                            & 0                                 & 6                           & 0                                & 1/5   \\
        MCTS+GR & \textbf{188}                & \textbf{125}                     & \textbf{179}                 & -                                 & \textbf{152}                & 25                               & 4/5   \\ \hline
        G       & \textbf{188}                & 41                               & \textbf{150}                 & 10                                & -                           & 0                                & 2/5   \\
        MCTS+G  & \textbf{188}                & \textbf{122}                     & \textbf{179}                 & \textbf{49}                       & \textbf{160}                & -                                & 5/5
    \end{tabular}
\end{table}

Table \ref{tab:mcts_greedy_diff} highlights the ranking of each method.
Unsurprisingly, the pure random heuristic is completely dominated by all methods.
The variant MCTS+Greedy is significantly better compared to the others.
In particular, it stays significantly better 49 times out of 188, versus 25 times in favor of the MCTS+Greedy-Random.
Indeed, it seems that for the WVCP, the greedy procedure, forcing the heaviest vertices to be grouped in the first colors enables a better organization of the color groups.
This is particularly true for the largest instances, where choosing random moves in the set of all legal moves is not very efficient as the number of color groups grows rapidly.
It explains also why the variant MCTS+Random performs badly on larger or denser instances such as the rxx instances or some difficult DIMACS instances.

However, with the deterministic simulation of the MCTS+Greedy variant, there is no sampling of the legal moves like in the MCTS+Greedy-Random variant allowing greater exploration of the search space and a better estimation of the most promising branches of the search tree.
This particularity of the MCTS+Greedy-Random variant allows us to find the BKS for more instances (see Table \ref{tab:mcts_greedy}).
Moreover, when exploring the results in the Appendix \ref{app:greedy}, one can see that the R75\_1gb instance from the DIMACS\_easy set is proved optimal by MCTS+Greedy-Random but not by MCTS+Greedy.
With stochastic help, the MCTS+Greedy-Random version can reach the best know score of 70, which leads to an early pruning of the tree that allows proving the optimality earlier.

\subsection{Exploitation vs exploration coefficient analysis}
\label{sec:coeff_c}

One key element of the MCTS algorithm is the coefficient $c$ balancing the compromise between exploration and exploitation in equation (\ref{eq:score_UCT}).
In this subsection, we investigate the importance of this coefficient by varying it and presenting the evolution of the score over time.
For this experimentation, we varied the coefficient $c$ from 0 (no exploration) to 5 (encourage exploration).
For each coefficient value, we performed 20 runs of the MCTS+Greedy-Random variant per instance during 5h per run (15h for the very large C2000.x instances)\footnote{This longer execution time explains some differences with the sensitivity analysis of this parameter made in \cite{GrelierGH22} with only 1h of computation time.}.

\begin{figure}[!h]
    \centering
    \includegraphics[width=0.45\textwidth,keepaspectratio]{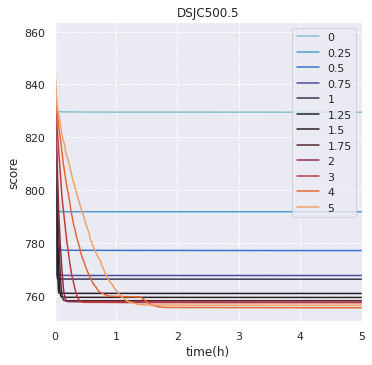}
    \includegraphics[width=0.45\textwidth,keepaspectratio]{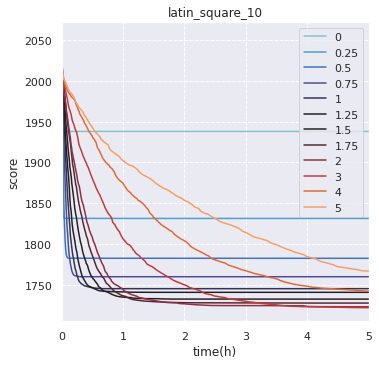}
    \includegraphics[width=0.45\textwidth,keepaspectratio]{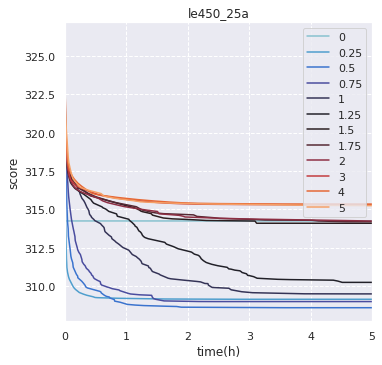}
    \includegraphics[width=0.45\textwidth,keepaspectratio]{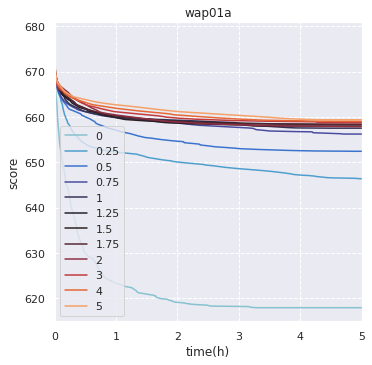}
    \includegraphics[width=0.45\textwidth,keepaspectratio]{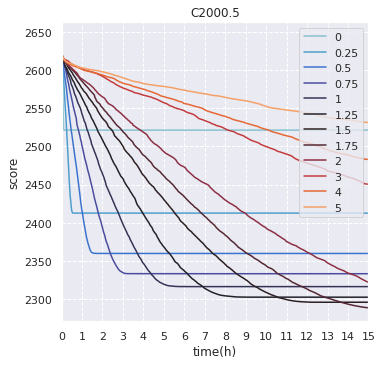}
    \includegraphics[width=0.45\textwidth,keepaspectratio]{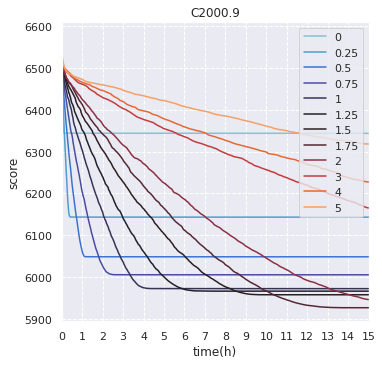}
    \caption{
        Plots of the evolution of the means of the scores over time for different values of the coefficient $c$ between 0 and 5, for the instances DSJC500.5, latin\_square\_10, le450\_25a, wap01a, C2000.5, and C2000.9.
        For each configuration, 20 runs are launched with the MCTS+Greedy-Random variant for 5h and 15h for the C2000 instances.
    }
    \label{fig:coeff}
\end{figure}

Figure \ref{fig:coeff} displays 6 plots showing the evolution of the mean of the best scores over time for the instances DSJC500.5, latin\_square\_10, le450\_25a, wap01a, C2000.5 and C2000.9 for the different values of the coefficient $c$.
These 6 instances come from the set of DIMACS\_hard instances and can be considered as very difficult.

Four typical patterns also seen for other instances are observed:

\begin{itemize}
    \item P1: instances requiring a lot of exploration,
    \item P2: instances requiring more exploration than exploitation,
    \item P3: instances requiring more exploitation than exploration,
    \item P4: instances requiring a lot of exploitation.
\end{itemize}

The first pattern P1 is observed for the instance DSJC500.5 and also queen instances.
For these instances, the lack of exploration leads to poor results, and better results are reached as the coefficient $c$ increases.
The pattern P2 is observed for the instance latin\_square\_10, but also for other instances such as flat1000 where the best score obtained in function of the coefficient $c$ has a U-shape, with an optimal value of $c$ between 1 and 2.
This phenomenon can also be observed on instances such as C2000.5 and C2000.9 where it becomes quickly more interesting to explore up to a certain point.
The pattern P3 found for the le450 instance shows the best results when there is only weak exploration, but the results are worse when $c$ is set to zero.

In general, for the patterns P1, P2 and P3, having no exploration at all rapidly leads to a local minimum trap and it seems better to secure a minimum of diversity to reach a better score, while, for the pattern P4, found for the wap instances (very large instances), giving a chance to the exploration leads the algorithm to be lost in the search space.
For very large instances, as the search tree is huge and cannot be sufficiently explored due to the time limit, it seems more beneficial for the algorithm to favor more intensification to better search for a good solution in a small part of the tree.
To sum, the most suitable exploration vs exploitation coefficient thus depends on the instance considered.
Finding the right general coefficient is a challenging task.
In this work, We adopted the coefficient $c=1$ for all other experiments.

\subsection{Monte Carlo Tree Search with local search}

This section studies the effects of the combination of MCTS with a local search heuristic.
Table \ref{tab:mcts_ls} summarizes the results of the local search procedures presented in Section \ref{sec:mcts_improved} followed by the combination of MCTS with each of these local search procedures during the simulation phase.
Each line presents the number of times a BKS is reached with the method.
Note that the objective of these MCTS variants coupled with a local search procedure is not to prove optimality.
For instances where the optimal score is known, MCTS and local search stop when this score is reached.

\begin{table}[h!]
    \centering
    \caption{
        Summary of the number of times the Best Known Score is reached for the local search procedures and MCTS combined with the different local search procedures.
        Values in bold highlight the best results for each line.
    }
    \label{tab:mcts_ls}
    \resizebox{\textwidth}{!}{
        \begin{tabular}{cc|cccccccc}
            instances    & \textbar I\textbar & AFISA & \multicolumn{1}{c|}{MCTS+AFISA}  & TW & \multicolumn{1}{c|}{MCTS+TW} & RedLS & \multicolumn{1}{c|}{MCTS+RedLS}   & ILSTS       & MCTS+ILSTS  \\ \hline
            pxx          & 35*                & 34    & \multicolumn{1}{c|}{34}          & 26 & \multicolumn{1}{c|}{34}      & 29    & \multicolumn{1}{c|}{\textbf{35}}  & \textbf{35} & \textbf{35} \\
            rxx          & 30*                & 5     & \multicolumn{1}{c|}{6}           & 7  & \multicolumn{1}{c|}{17}      & 7     & \multicolumn{1}{c|}{26}           & \textbf{30} & \textbf{30} \\
            DIMACS\_easy & 59*                & 55    & \multicolumn{1}{c|}{\textbf{58}} & 42 & \multicolumn{1}{c|}{54}      & 54    & \multicolumn{1}{c|}{\textbf{58}}  & \textbf{58} & \textbf{58} \\
            DIMACS\_hard & 64                 & 18    & \multicolumn{1}{c|}{22}          & 20 & \multicolumn{1}{c|}{23}      & 28    & \multicolumn{1}{c|}{\textbf{34}}  & 27          & 27          \\ \hline
            Total        & 188                & 112   & \multicolumn{1}{c|}{120}         & 95 & \multicolumn{1}{c|}{128}     & 118   & \multicolumn{1}{c|}{\textbf{153}} & 150         & 150
        \end{tabular}
    }
\end{table}

We observe from this table that the local search procedure allowing to reach the highest number of BKS is ILTS, followed by RedLS. TW, AFISA, and RedLS also improve their results when coupled with the MCTS framework.
In particular, the variant MCTS+RedlLS finds the BKS for 34 difficult instances of the DIMACS\_hard set and 153 instances over 188 in total, which is better than all other algorithms tested in this study.
Moreover, it is worth noting that this combination allowed to find two new records for the difficult DIMACS instances le450\_15a and queen12\_12gb that were never reported in the literature
(see detailed results in Tables \ref{tab:app11} and \ref{tab:app12}).
It highlights that the MCTS framework proposed in this paper can help a local search algorithm such as RedLS to continuously find new promising starting points in the search space.

However, when coupled with the ILSTS algorithm (variant MCTS+ILTS), it does not seem to improve the results.
It may be explained by the fact that ILSTS is an iterated local search algorithm already integrating various perturbation mechanisms allowing to escape local optima.
Therefore, finding new good starting points in the search space guided by MCTS seems less interesting for ILSTS than for other local search procedures such as TW, AFISA, and RedLS.

Table \ref{tab:mcts_ls_diff} shows pairwise comparisons between all the MCTS variants and all the local search procedures.

\begin{table}[h!]
    \centering
    \caption{
        Comparison between all the MCTS (MC) variants and all the local search procedures.
        Each row corresponds to the number of times the method of the row is significantly better than the method on the column on the 188 instances. NOT CLEAR, REPHRASE
        For example, out of 188 instances, TW is better than MCTS+TW on 24 instances, 43 times against RedLS and 5 times against MCTS+RedLS.
        Values in bold means that method $a$ is more often better than method $b$ when we look at how many time $b$ is significantly better than $a$.
        The Total column corresponds to the number of times a method is in bold on the line.
    }
    \label{tab:mcts_ls_diff}
    \begin{tabular}{c|cc|cc|cc|cc|cc|c}
                   & \rotatebox[origin=c]{90}{MCTS+GR} & \rotatebox[origin=c]{90}{MCTS+G} & \rotatebox[origin=c]{90}{AFISA} & \rotatebox[origin=c]{90}{MCTS+AFISA} & \rotatebox[origin=c]{90}{TW} & \rotatebox[origin=c]{90}{MCTS+TW} & \rotatebox[origin=c]{90}{RedLS} & \rotatebox[origin=c]{90}{MCTS+RedLS} & \rotatebox[origin=c]{90}{ILSTS} & \rotatebox[origin=c]{90}{MCTS+ILSTS} & Total \\ \hline
        MCTS+GR    & -                                 & 25                               & 43                              & 30                                   & 50                           & 15                                & 55                              & 2                                    & 7                               & 12                                   & 0/9   \\
        MCTS+G     & \textbf{49}                       & -                                & \textbf{54}                     & \textbf{48}                          & \textbf{60}                  & 28                                & 56                              & 10                                   & 16                              & 24                                   & 4/9   \\\hline
        AFISA      & \textbf{45}                       & 35                               & -                               & 22                                   & 33                           & 14                                & 34                              & 1                                    & 12                              & 18                                   & 1/9   \\
        MCTS+AFISA & \textbf{56}                       & 47                               & \textbf{62}                     & -                                    & \textbf{62}                  & 6                                 & \textbf{66}                     & 3                                    & 1                               & 8                                    & 4/9   \\\hline
        TW         & \textbf{60}                       & 55                               & \textbf{56}                     & 40                                   & -                            & 24                                & 43                              & 5                                    & 18                              & 27                                   & 2/9   \\
        MCTS+TW    & \textbf{74}                       & \textbf{62}                      & \textbf{76}                     & \textbf{64}                          & \textbf{73}                  & -                                 & \textbf{76}                     & 7                                    & 13                              & 25                                   & 6/9   \\\hline
        RedLS      & \textbf{62}                       & \textbf{68}                      & \textbf{62}                     & 43                                   & \textbf{61}                  & 38                                & -                               & 15                                   & 28                              & 33                                   & 4/9   \\
        MCTS+RedLS & \textbf{95}                       & \textbf{94}                      & \textbf{103}                    & \textbf{73}                          & \textbf{104}                 & \textbf{65}                       & \textbf{97}                     & -                                    & \textbf{39}                     & \textbf{43}                          & 9/9   \\\hline
        ILSTS      & \textbf{90}                       & \textbf{79}                      & \textbf{88}                     & \textbf{71}                          & \textbf{83}                  & \textbf{45}                       & \textbf{81}                     & 13                                   & -                               & \textbf{37}                          & 8/9   \\
        MCTS+ILSTS & \textbf{84}                       & \textbf{77}                      & \textbf{80}                     & \textbf{62}                          & \textbf{74}                  & \textbf{36}                       & \textbf{80}                     & 13                                   & 0                               & -                                    & 7/9
    \end{tabular}
\end{table}

When comparing the performances of the local search algorithms by looking at the number of times they are significantly better than the others in terms of average score, AFISA and TW switch their ranks, and RedLS and ILSTS keep the second and first place.

When comparing MCTS+Greedy-Random and MCTS+Greedy to the local search or MCTS+local\_search variants,
MCTS+Greedy-Random does not show much success compared to MCTS+Greedy, which has better results more often against AFISA, MCTS+AFISA, and TW for a few more instances.

The MCTS+local\_search variants always improve the number of BKS found and are more often significantly better than the corresponding local search only, except for ILSTS, which is better on its own and dominates the MCTS+ILSTS variant.

The variant MCTS+RedLS is significantly better than the other methods, even compared to ILSTS only.
These results indicate that combining MCTS with a local search is of interest to improve the underlying local search procedures such as AFISA, TW, and RedLS.

%%%%%%%%%%%%%%%%%%%%%%%%%%%%%%%%%%%%%%%%%%%%%%%%%%%%%%%%%%%%%%%%%%%%%%%%%%%%%%%%%%%%%%%%%%%%%%%%%%%%
%                                            Conclusion
%%%%%%%%%%%%%%%%%%%%%%%%%%%%%%%%%%%%%%%%%%%%%%%%%%%%%%%%%%%%%%%%%%%%%%%%%%%%%%%%%%%%%%%%%%%%%%%%%%%%

\section{Conclusions} \label{sec:conclusion}

In this work, we investigated Monte Carlo Tree Search applied to the weighted vertex coloring problem.
We studied different greedy strategies and local searches used for the simulation phase.
Our experimental results lead to three conclusions.

When the instance is large and when a time limit is imposed, MCTS does not have the time to learn promising areas in the search space and it seems more beneficial to favor more intensification, which can be done in three different ways: (i) by lowering the coefficient which balances the compromise between exploitation and exploration during the selection phase, (ii) by using a dedicated heuristic exploiting the specificity of the problem (grouping in priority the heaviest vertices in the first groups of colors), (iii) and by using a local search procedure to improve the complete solution.

Conversely, for small instances, it seems more beneficial to encourage more exploration, to avoid getting stuck in local optima.
It can be done, by increasing the coefficient, which balances the compromise between exploitation and exploration, and by using a simulation strategy with more randomness, which favors more exploration of the search tree and also allows a better evaluation of the most promising branches of the MCTS tree.
For these small instances, the MCTS algorithm can provide some optimality proofs.

For medium instances, it seems important to find a good compromise between exploration and exploitation.
For such instances, coupling the MCTS algorithm with a local search procedure allows finding better solutions, which cannot be reached by the MCTS algorithm or the local search alone.

Other future works could be envisaged.
For example, an interesting study would be to automatically choose the balance coefficient between exploitation and exploration on the fly when solving each specific instance.
It could also be interesting to use a more adaptive approach to trigger the local search, or to use a machine-learning algorithm to guide the search toward more promising branches of the search tree.

\section*{Acknowledgements} \label{sec:acknowledgements}

We would like to thank Dr. Wen Sun \cite{Sun_Hao_Lai_Wu_2018}, Dr. Yiyuan Wang, \cite{Wang_Cai_Pan_Li_Yin_2020} and Pr. Bruno Nogueira \cite{Nogueira_Tavares_Maciel_2021} for sharing their codes.
We also thank the organizers and editors of Evostar 2022 to give us the opportunity to present this extended version of our conference paper \cite{GrelierGH22} for this Topical Issue of the SN Computer Science journal.
This work was granted access to the HPC resources of IDRIS (Grant No. 2020-A0090611887) from GENCI and the Centre Régional de Calcul Intensif des Pays de la Loire (CCIPL).

\section*{Authors' contributions}

C. Grelier developed the code, prepared the reduced instances, and performed the tests. O. Goudet and J.K. Hao planned and supervised the work.
All the authors contributed to the analysis and the writing of the manuscript.

\section*{Conflict of Interest}

The authors declare that they have no conflicts of interest.

\section*{Availability of data set, code, and results}

The instances used in this work are available at \url{https://github.com/Cyril-Grelier/gc_instances}.
The codes of the tested methods and their detailed results are available at \url{https://github.com/Cyril-Grelier/gc_wvcp_mcts}.

\bibliography{biblio}


\begin{thebibliography}{25}
\ifx \bisbn   \undefined \def \bisbn  #1{ISBN #1}\fi
\ifx \binits  \undefined \def \binits#1{#1}\fi
\ifx \bauthor  \undefined \def \bauthor#1{#1}\fi
\ifx \batitle  \undefined \def \batitle#1{#1}\fi
\ifx \bjtitle  \undefined \def \bjtitle#1{#1}\fi
\ifx \bvolume  \undefined \def \bvolume#1{\textbf{#1}}\fi
\ifx \byear  \undefined \def \byear#1{#1}\fi
\ifx \bissue  \undefined \def \bissue#1{#1}\fi
\ifx \bfpage  \undefined \def \bfpage#1{#1}\fi
\ifx \blpage  \undefined \def \blpage #1{#1}\fi
\ifx \burl  \undefined \def \burl#1{\textsf{#1}}\fi
\ifx \doiurl  \undefined \def \doiurl#1{\url{https://doi.org/#1}}\fi
\ifx \betal  \undefined \def \betal{\textit{et al.}}\fi
\ifx \binstitute  \undefined \def \binstitute#1{#1}\fi
\ifx \binstitutionaled  \undefined \def \binstitutionaled#1{#1}\fi
\ifx \bctitle  \undefined \def \bctitle#1{#1}\fi
\ifx \beditor  \undefined \def \beditor#1{#1}\fi
\ifx \bpublisher  \undefined \def \bpublisher#1{#1}\fi
\ifx \bbtitle  \undefined \def \bbtitle#1{#1}\fi
\ifx \bedition  \undefined \def \bedition#1{#1}\fi
\ifx \bseriesno  \undefined \def \bseriesno#1{#1}\fi
\ifx \blocation  \undefined \def \blocation#1{#1}\fi
\ifx \bsertitle  \undefined \def \bsertitle#1{#1}\fi
\ifx \bsnm \undefined \def \bsnm#1{#1}\fi
\ifx \bsuffix \undefined \def \bsuffix#1{#1}\fi
\ifx \bparticle \undefined \def \bparticle#1{#1}\fi
\ifx \barticle \undefined \def \barticle#1{#1}\fi
\bibcommenthead
\ifx \bconfdate \undefined \def \bconfdate #1{#1}\fi
\ifx \botherref \undefined \def \botherref #1{#1}\fi
\ifx \url \undefined \def \url#1{\textsf{#1}}\fi
\ifx \bchapter \undefined \def \bchapter#1{#1}\fi
\ifx \bbook \undefined \def \bbook#1{#1}\fi
\ifx \bcomment \undefined \def \bcomment#1{#1}\fi
\ifx \oauthor \undefined \def \oauthor#1{#1}\fi
\ifx \citeauthoryear \undefined \def \citeauthoryear#1{#1}\fi
\ifx \endbibitem  \undefined \def \endbibitem {}\fi
\ifx \bconflocation  \undefined \def \bconflocation#1{#1}\fi
\ifx \arxivurl  \undefined \def \arxivurl#1{\textsf{#1}}\fi
\csname PreBibitemsHook\endcsname

%%% 1
\bibitem{GrelierGH22}
\begin{bchapter}
\bauthor{\bsnm{Grelier}, \binits{C.}},
\bauthor{\bsnm{Goudet}, \binits{O.}},
\bauthor{\bsnm{Hao}, \binits{J.-K.}}:
\bctitle{On monte carlo tree search for weighted vertex coloring}.
In: \beditor{\bsnm{Pérez~Cáceres}, \binits{L.}},
\beditor{\bsnm{Verel}, \binits{S.}} (eds.)
\bbtitle{Evolutionary Computation in Combinatorial Optimization}.
\bsertitle{Lecture Notes in Computer Science},
vol. \bseriesno{13222},
pp. \bfpage{1}--\blpage{16}
(\byear{2022})
\end{bchapter}
\endbibitem

%%% 2
\bibitem{lewis_guide_2021}
\begin{bbook}
\bauthor{\bsnm{Lewis}, \binits{R.}}:
\bbtitle{Guide to Graph Colouring: Algorithms and Applications. Springer},
(\byear{2021})
\end{bbook}
\endbibitem

%%% 3
\bibitem{Goudet_Grelier_Hao_2021}
\begin{botherref}
\oauthor{\bsnm{Goudet}, \binits{O.}},
\oauthor{\bsnm{Grelier}, \binits{C.}},
\oauthor{\bsnm{Hao}, \binits{J.-K.}}:
A deep learning guided memetic framework for graph coloring problems.
arXiv:2109.05948 [cs]
(2021)
\end{botherref}
\endbibitem

%%% 4
\bibitem{Nogueira_Tavares_Maciel_2021}
\begin{barticle}
\bauthor{\bsnm{Nogueira}, \binits{B.}},
\bauthor{\bsnm{Tavares}, \binits{E.}},
\bauthor{\bsnm{Maciel}, \binits{P.}}:
\batitle{Iterated local search with tabu search for the weighted vertex
  coloring problem}.
\bjtitle{Computers \& Operations Research}
\bvolume{125},
\bfpage{105087}
(\byear{2021})
\end{barticle}
\endbibitem

%%% 5
\bibitem{Sun_Hao_Lai_Wu_2018}
\begin{barticle}
\bauthor{\bsnm{Sun}, \binits{W.}},
\bauthor{\bsnm{Hao}, \binits{J.-K.}},
\bauthor{\bsnm{Lai}, \binits{X.}},
\bauthor{\bsnm{Wu}, \binits{Q.}}:
\batitle{Adaptive feasible and infeasible tabu search for weighted vertex
  coloring}.
\bjtitle{Information Sciences}
\bvolume{466},
\bfpage{203}--\blpage{219}
(\byear{2018})
\end{barticle}
\endbibitem

%%% 6
\bibitem{Wang_Cai_Pan_Li_Yin_2020}
\begin{bchapter}
\bauthor{\bsnm{Wang}, \binits{S.} \bsuffix{Yiyuanand~Cai}},
\bauthor{\bsnm{Pan}, \binits{S.}},
\bauthor{\bsnm{Li}, \binits{X.}},
\bauthor{\bsnm{Yin}, \binits{M.}}:
\bctitle{Reduction and local search for weighted graph coloring problem}.
In: \bbtitle{The Thirty-Fourth {AAAI} Conference on Artificial Intelligence,
  {AAAI} 2020, New York, NY, USA, February 7-12, 2020},
pp. \bfpage{2433}--\blpage{2441}
(\byear{2020})
\end{bchapter}
\endbibitem

%%% 7
\bibitem{Prais_Ribeiro_2000}
\begin{barticle}
\bauthor{\bsnm{Prais}, \binits{M.}},
\bauthor{\bsnm{Ribeiro}, \binits{C.C.}}:
\batitle{Reactive grasp: An application to a matrix decomposition problem in
  tdma traffic assignment}.
\bjtitle{INFORMS Journal on Computing}
\bvolume{12}(\bissue{3}),
\bfpage{164}--\blpage{176}
(\byear{2000})
\end{barticle}
\endbibitem

%%% 8
\bibitem{Pemmaraju_Raman_2005}
\begin{bchapter}
\bauthor{\bsnm{Pemmaraju}, \binits{S.V.}},
\bauthor{\bsnm{Raman}, \binits{R.}}:
\bctitle{Approximation algorithms for the max-coloring problem}.
In: \beditor{\bsnm{Caires}, \binits{L.}},
\beditor{\bsnm{Italiano}, \binits{G.F.}},
\beditor{\bsnm{Monteiro}, \binits{L.}},
\beditor{\bsnm{Palamidessi}, \binits{C.}},
\beditor{\bsnm{Yung}, \binits{M.}} (eds.)
\bbtitle{Automata, Languages and Programming}.
\bsertitle{Lecture Notes in Computer Science},
pp. \bfpage{1064}--\blpage{1075}
(\byear{2005})
\end{bchapter}
\endbibitem

%%% 9
\bibitem{furini2012exact}
\begin{barticle}
\bauthor{\bsnm{Furini}, \binits{F.}},
\bauthor{\bsnm{Malaguti}, \binits{E.}}:
\batitle{Exact weighted vertex coloring via branch-and-price}.
\bjtitle{Discrete Optimization}
\bvolume{9}(\bissue{2}),
\bfpage{130}--\blpage{136}
(\byear{2012})
\end{barticle}
\endbibitem

%%% 10
\bibitem{malaguti2009models}
\begin{barticle}
\bauthor{\bsnm{Malaguti}, \binits{E.}},
\bauthor{\bsnm{Monaci}, \binits{M.}},
\bauthor{\bsnm{Toth}, \binits{P.}}:
\batitle{Models and heuristic algorithms for a weighted vertex coloring
  problem}.
\bjtitle{Journal of Heuristics}
\bvolume{15}(\bissue{5}),
\bfpage{503}--\blpage{526}
(\byear{2009})
\end{barticle}
\endbibitem

%%% 11
\bibitem{Cornaz_Furini_Malaguti_2017}
\begin{barticle}
\bauthor{\bsnm{Cornaz}, \binits{D.}},
\bauthor{\bsnm{Furini}, \binits{F.}},
\bauthor{\bsnm{Malaguti}, \binits{E.}}:
\batitle{Solving vertex coloring problems as maximum weight stable set
  problems}.
\bjtitle{Discrete Applied Mathematics}
\bvolume{217},
\bfpage{151}--\blpage{162}
(\byear{2017})
\end{barticle}
\endbibitem

%%% 12
\bibitem{zhou2018improving}
\begin{barticle}
\bauthor{\bsnm{Zhou}, \binits{Y.}},
\bauthor{\bsnm{Duval}, \binits{B.}},
\bauthor{\bsnm{Hao}, \binits{J.-K.}}:
\batitle{Improving probability learning based local search for graph coloring}.
\bjtitle{Applied Soft Computing}
\bvolume{65},
\bfpage{542}--\blpage{553}
(\byear{2018})
\end{barticle}
\endbibitem

%%% 13
\bibitem{zhou2020frequent}
\begin{barticle}
\bauthor{\bsnm{Zhou}, \binits{Y.}},
\bauthor{\bsnm{Hao}, \binits{J.-K.}},
\bauthor{\bsnm{Duval}, \binits{B.}}:
\batitle{Frequent pattern-based search: A case study on the quadratic
  assignment problem}.
\bjtitle{IEEE Transactions on Systems, Man, and Cybernetics: Systems}
\bvolume{52}(\bissue{3}),
\bfpage{1503}--\blpage{1515}
(\byear{2022})
\end{barticle}
\endbibitem

%%% 14
\bibitem{gelly2006modification}
\begin{botherref}
\oauthor{\bsnm{Gelly}, \binits{S.}},
\oauthor{\bsnm{Wang}, \binits{Y.}},
\oauthor{\bsnm{Munos}, \binits{R.}},
\oauthor{\bsnm{Teytaud}, \binits{O.}}:
Modification of uct with patterns in monte-carlo go.
PhD thesis,
INRIA
(2006)
\end{botherref}
\endbibitem

%%% 15
\bibitem{browne2012survey}
\begin{barticle}
\bauthor{\bsnm{Browne}, \binits{C.B.}},
\bauthor{\bsnm{Powley}, \binits{E.}},
\bauthor{\bsnm{Whitehouse}, \binits{D.}},
\bauthor{\bsnm{Lucas}, \binits{S.M.}},
\bauthor{\bsnm{Cowling}, \binits{P.I.}},
\bauthor{\bsnm{Rohlfshagen}, \binits{P.}},
\bauthor{\bsnm{Tavener}, \binits{S.}},
\bauthor{\bsnm{Perez}, \binits{D.}},
\bauthor{\bsnm{Samothrakis}, \binits{S.}},
\bauthor{\bsnm{Colton}, \binits{S.}}:
\batitle{A survey of monte carlo tree search methods}.
\bjtitle{IEEE Transactions on Computational Intelligence and AI in Games}
\bvolume{4}(\bissue{1}),
\bfpage{1}--\blpage{43}
(\byear{2012})
\end{barticle}
\endbibitem

%%% 16
\bibitem{Silver_Huang_Maddison_2016}
\begin{barticle}
\bauthor{\bsnm{Silver}, \binits{D.}},
\bauthor{\bsnm{Huang}, \binits{A.}},
\bauthor{\bsnm{Maddison}, \binits{C.J.}},
\bauthor{\bsnm{Guez}, \binits{A.}},
\bauthor{\bsnm{Sifre}, \binits{L.}},
\bauthor{\bparticle{van~den} \bsnm{Driessche}, \binits{G.}},
\bauthor{\bsnm{Schrittwieser}, \binits{J.}},
\bauthor{\bsnm{Antonoglou}, \binits{I.}},
\bauthor{\bsnm{Panneershelvam}, \binits{V.}},
\bauthor{\bsnm{Lanctot}, \binits{M.}},
\bauthor{\bparticle{et} \bsnm{al.}}:
\batitle{Mastering the game of go with deep neural networks and tree search}.
\bjtitle{Nature}
\bvolume{529}(\bissue{7587}),
\bfpage{484}--\blpage{489}
(\byear{2016})
\end{barticle}
\endbibitem

%%% 17
\bibitem{edelkamp2014solving}
\begin{bchapter}
\bauthor{\bsnm{Edelkamp}, \binits{S.}},
\bauthor{\bsnm{Greulich}, \binits{C.}}:
\bctitle{Solving physical traveling salesman problems with policy adaptation}.
In: \bbtitle{2014 IEEE Conference on Computational Intelligence and Games},
pp. \bfpage{1}--\blpage{8}
(\byear{2014}).
\bcomment{IEEE}
\end{bchapter}
\endbibitem

%%% 18
\bibitem{Jooken_Leyman_2020}
\begin{botherref}
\oauthor{\bsnm{Jooken}, \binits{J.}},
\oauthor{\bsnm{Leyman}, \binits{P.}},
\oauthor{\bsnm{De~Causmaecker}, \binits{P.}},
\oauthor{\bsnm{Wauters}, \binits{T.}}:
Exploring search space trees using an adapted version of monte carlo tree
  search for combinatorial optimization problems.
arXiv:2010.11523 [cs, math]
(2020)
\end{botherref}
\endbibitem

%%% 19
\bibitem{Cazenave_Negrevergne_Sikora_2020}
\begin{bchapter}
\bauthor{\bsnm{Cazenave}, \binits{T.}},
\bauthor{\bsnm{Negrevergne}, \binits{B.}},
\bauthor{\bsnm{Sikora}, \binits{F.}}:
\bctitle{Monte carlo graph coloring}.
In: \bbtitle{Monte Carlo Search 2020, IJCAI Workshop}
(\byear{2020})
\end{bchapter}
\endbibitem

%%% 20
\bibitem{brelaz1979new}
\begin{barticle}
\bauthor{\bsnm{Br{\'e}laz}, \binits{D.}}:
\batitle{New methods to color the vertices of a graph}.
\bjtitle{Communications of the ACM}
\bvolume{22}(\bissue{4}),
\bfpage{251}--\blpage{256}
(\byear{1979})
\end{barticle}
\endbibitem

%%% 21
\bibitem{KubaleJ84}
\begin{barticle}
\bauthor{\bsnm{Kubale}, \binits{M.}},
\bauthor{\bsnm{Jackowski}, \binits{B.}}:
\batitle{A generalized implicit enumeration algorithm for graph coloring}.
\bjtitle{Communications of the {ACM}}
\bvolume{28}(\bissue{4}),
\bfpage{412}--\blpage{418}
(\byear{1985})
\end{barticle}
\endbibitem

%%% 22
\bibitem{Lai_Robbins_1985}
\begin{barticle}
\bauthor{\bsnm{Lai}, \binits{T.L.}},
\bauthor{\bsnm{Robbins}, \binits{H.}}:
\batitle{Asymptotically efficient adaptive allocation rules}.
\bjtitle{Advances in Applied Mathematics}
\bvolume{6}(\bissue{1}),
\bfpage{4}--\blpage{22}
(\byear{1985})
\end{barticle}
\endbibitem

%%% 23
\bibitem{Hertz_Werra_1987}
\begin{barticle}
\bauthor{\bsnm{Hertz}, \binits{A.}},
\bauthor{\bparticle{de} \bsnm{Werra}, \binits{D.}}:
\batitle{Using tabu search techniques for graph coloring}.
\bjtitle{Computing}
\bvolume{39},
\bfpage{345}--\blpage{351}
(\byear{1987})
\end{barticle}
\endbibitem

%%% 24
\bibitem{cai_local_2011}
\begin{barticle}
\bauthor{\bsnm{Cai}, \binits{S.}},
\bauthor{\bsnm{Su}, \binits{K.}},
\bauthor{\bsnm{Sattar}, \binits{A.}}:
\batitle{Local search with edge weighting and configuration checking heuristics
  for minimum vertex cover}.
\bjtitle{Artificial Intelligence}
\bvolume{175}(\bissue{9}),
\bfpage{1672}--\blpage{1696}
(\byear{2011})
\end{barticle}
\endbibitem

%%% 25
\bibitem{Cheeseman_Kanefsky_Taylor_1991}
\begin{bchapter}
\bauthor{\bsnm{Cheeseman}, \binits{P.}},
\bauthor{\bsnm{Kanefsky}, \binits{B.}},
\bauthor{\bsnm{Taylor}, \binits{W.M.}}:
\bctitle{Where the really hard problems are}.
In: \bbtitle{Proceedings of the 12th International Joint Conference on
  Artificial Intelligence - Volume 1},
pp. \bfpage{331}--\blpage{337}
(\byear{1991})
\end{bchapter}
\endbibitem

\end{thebibliography}
% \bibliography{main.bbl}

%%%%%%%%%%%%%%%%%%%%%%%%%%%%%%%%%%%%%%%%%%%%%%%%%%%%%%%%%%%%%%%%%%%%%%%%%%%%%%%%%%%%%%%%%%%%%%%%%%%%
%                                            Appendix
%%%%%%%%%%%%%%%%%%%%%%%%%%%%%%%%%%%%%%%%%%%%%%%%%%%%%%%%%%%%%%%%%%%%%%%%%%%%%%%%%%%%%%%%%%%%%%%%%%%%

\appendix

\section{Appendix}

In this Appendix, we present the detailed results on each instance of the different algorithms presented in this paper.
In each of the following tables, the first column is the instance name followed by the number of vertices, the number of edges, and the best-known scores of the literature.
Then for each method, are displayed the best score, the average score (out of 20 runs), and the average execution time in seconds required to reach the best score.

Best-known scores that are proved optimal are indicated with a star (*) next to the score.
Scores in bold means that the score is equal to the BKS, except two times in Tables \ref{tab:app11} and \ref{tab:app12} when the proposed variant MCTS+RedLS improves the BKS on le450\_15a and queen12\_12gb (which are indicated with a plus (+)).
More readable spreadsheets of those tables (with some more information) can be found in the GitHub repository \url{https://github.com/Cyril-Grelier/gc\_instances} (follow the indications in the README file).

\subsection{Detailed results of MCTS coupled with greedy strategies}\label{app:greedy}

Tables \ref{tab:app1}- \ref{tab:app6} report the detailed results for the MCTS variants combined with the greedy strategies (cf. Section \ref{sec:MCTS_greedy}).

\begin{table}[!hp]
    \scriptsize
    \caption{Results on pxx instances for Greedy and MCTS+Greedy methods.}
    \label{tab:app1}
    \Rotatebox{90}{
        \resizebox{1.7\textwidth}{!}{
\begin{tabular}{cccc|ccc|ccc|ccc|ccc|ccc|ccc}
                \multirow{2}{*}{instance}          & \multirow{2}{*}{\textbar $V$\textbar} & \multirow{2}{*}{\textbar $E$\textbar} & \multirow{2}{*}{BKS}      & \multicolumn{3}{c|}{Random} & \multicolumn{3}{c|}{MCTS+Random} & \multicolumn{3}{c|}{Greedy-Random} & \multicolumn{3}{c|}{MCTS+Greedy-Random} & \multicolumn{3}{c|}{Greedy} & \multicolumn{3}{c}{MCTS+Greedy}                                                                                                                               \\
                                                   &                                       &                                       &                           & best                        & avg                              & time                               & best                                    & avg                         & time                            & best          & avg    & time & best           & avg    & time & best          & avg  & time & best           & avg  & time \\ \hline
                p06                                & 16                                    & 38                                    & 565*                      & 605                         & 853.9                            & 0                                  & \textbf{565*}                           & 565.2                       & 0                               & \textbf{565}  & 579    & 0    & \textbf{565*}  & 565    & 0    & 585           & 585  & 0    & \textbf{565*}  & 565  & 0    \\
\end{tabular}
        }
    }
\end{table}

\begin{table}[!hp]
    \scriptsize
    \caption{Results on rxx instances for Greedy and MCTS+Greedy methods.}
    \label{tab:app2}
    \Rotatebox{90}{
        \resizebox{1.7\textwidth}{!}{
            \begin{tabular}{cccc|ccc|ccc|ccc|ccc|ccc|ccc}
                \multirow{2}{*}{instance}          & \multirow{2}{*}{\textbar $V$\textbar} & \multirow{2}{*}{\textbar $E$\textbar} & \multirow{2}{*}{BKS}     & \multicolumn{3}{c|}{Random} & \multicolumn{3}{c|}{MCTS+Random} & \multicolumn{3}{c|}{Greedy-Random} & \multicolumn{3}{c|}{MCTS+Greedy-Random} & \multicolumn{3}{c|}{Greedy} & \multicolumn{3}{c}{MCTS+Greedy}                                                                                                                            \\
                                                   &                                       &                                       &                          & best                        & avg                              & time                               & best                                    & avg                         & time                            & best  & avg     & time & best          & avg    & time   & best          & avg  & time & best          & avg    & time   \\ \hline
                r01                                & 144                                   & 1280                                  & 6724*                    & 9956                        & 11532.3                          & 0                                  & 6745                                    & 6810.6                      & 11                              & 7023  & 7203.9  & 0    & \textbf{6724} & 6725.4 & 30.8   & 7138          & 7138 & 0    & \textbf{6724} & 6724   & 0      \\
\end{tabular}
        }
    }
\end{table}

\begin{table}[!hp]
    \scriptsize
    \caption{Results on DIMACS instances with proven optima for Greedy and MCTS+Greedy methods (1/2).}
    \label{tab:app3}
    \Rotatebox{90}{
        \resizebox{1.7\textwidth}{!}{
            \begin{tabular}{cccc|ccc|ccc|ccc|ccc|ccc|ccc}
                \multirow{2}{*}{instance} & \multirow{2}{*}{\textbar $V$\textbar} & \multirow{2}{*}{\textbar $E$\textbar} & \multirow{2}{*}{BKS} & \multicolumn{3}{c|}{Random} & \multicolumn{3}{c|}{MCTS+Random} & \multicolumn{3}{c|}{Greedy-Random} & \multicolumn{3}{c|}{MCTS+Greedy-Random} & \multicolumn{3}{c|}{Greedy} & \multicolumn{3}{c}{MCTS+Greedy}                                                                                                                          \\
                                          &                                       &                                       &                      & best                        & avg                              & time                               & best                                    & avg                         & time                            & best        & avg    & time & best         & avg   & time   & best        & avg  & time & best         & avg  & time   \\ \hline
                DSJC125.9g                & 125                                   & 6961                                  & 169*                 & 207                         & 217.8                            & 0                                  & 172                                     & 176.6                       & 2040                            & 180         & 186.4  & 0    & 171          & 172.7 & 2045.3 & 182         & 182  & 0    & 173          & 173  & 1      \\
                DSJC125.9gb               & 125                                   & 6961                                  & 604*                 & 732                         & 765.6                            & 0                                  & 614                                     & 631.9                       & 1640                            & 647         & 669.1  & 0    & 614          & 617.5 & 377.5  & 656         & 656  & 0    & 621          & 621  & 1      \\
\end{tabular}
        }
    }
\end{table}

\begin{table}[!hp]
    \scriptsize
    \caption{Results on DIMACS instances with proven optima for Greedy and MCTS+Greedy methods (2/2).}
    \label{tab:app4}
    \Rotatebox{90}{
        \resizebox{1.7\textwidth}{!}{
            \begin{tabular}{cccc|ccc|ccc|ccc|ccc|ccc|ccc}
                \multirow{2}{*}{instance}          & \multirow{2}{*}{\textbar $V$\textbar} & \multirow{2}{*}{\textbar $E$\textbar} & \multirow{2}{*}{BKS}      & \multicolumn{3}{c|}{Random} & \multicolumn{3}{c|}{MCTS+Random} & \multicolumn{3}{c|}{Greedy-Random} & \multicolumn{3}{c|}{MCTS+Greedy-Random} & \multicolumn{3}{c|}{Greedy} & \multicolumn{3}{c}{MCTS+Greedy}                                                                                                                             \\
                                                   &                                       &                                       &                           & best                        & avg                              & time                               & best                                    & avg                         & time                            & best         & avg   & time & best          & avg   & time   & best         & avg & time & best          & avg   & time   \\ \hline
                R100\_1g                           & 99                                    & 506                                   & 21*                       & 46                          & 58.5                             & 0                                  & 24                                      & 25.6                        & 703.2                           & 26           & 28.4  & 0    & 22            & 22.7  & 1743.5 & 26           & 26  & 0    & 23            & 23    & 0      \\
\end{tabular}
        }
    }
\end{table}

\begin{table}[!hp]
    \scriptsize
    \caption{Results on DIMACS instances with unknown optima for Greedy and MCTS+Greedy methods (1/2).}
    \label{tab:app5}
    \Rotatebox{90}{
        \resizebox{1.7\textwidth}{!}{
            \begin{tabular}{cccc|ccc|ccc|ccc|ccc|ccc|ccc}
                \multirow{2}{*}{instance} & \multirow{2}{*}{\textbar $V$\textbar} & \multirow{2}{*}{\textbar $E$\textbar} & \multirow{2}{*}{BKS} & \multicolumn{3}{c|}{Random} & \multicolumn{3}{c|}{MCTS+Random} & \multicolumn{3}{c|}{Greedy-Random} & \multicolumn{3}{c|}{MCTS+Greedy-Random} & \multicolumn{3}{c|}{Greedy} & \multicolumn{3}{c}{MCTS+Greedy}                                                                                                                               \\
                                          &                                       &                                       &                      & best                        & avg                              & time                               & best                                    & avg                         & time                            & best         & avg    & time & best         & avg    & time   & best         & avg  & time & best         & avg    & time   \\ \hline
                C2000.5                   & 2000                                  & 999836                                & 2151                 & 3401                        & 3443                             & 0                                  & 3184                                    & 3210                        & 3528                            & 2655         & 2689.7 & 0    & 2523         & 2543.6 & 3436   & 2470         & 2470 & 0    & 2368         & 2371.5 & 3326.8 \\
                C2000.9                   & 2000                                  & 1799532                               & 5486                 & 7465                        & 7553.8                           & 0                                  & 7066                                    & 7175.7                      & 3446                            & 6568         & 6648.3 & 0    & 6276         & 6305   & 3455   & 6449         & 6449 & 0    & 6184         & 6192.9 & 3528   \\
\end{tabular}
        }
    }
\end{table}

\begin{table}[!hp]
    \scriptsize
    \caption{Results on DIMACS instances with unknown optima for Greedy and MCTS+Greedy methods (2/2).}
    \label{tab:app6}
    \Rotatebox{90}{
        \resizebox{1.7\textwidth}{!}{
            \begin{tabular}{cccc|ccc|ccc|ccc|ccc|ccc|ccc}
                \multirow{2}{*}{instance}          & \multirow{2}{*}{\textbar $V$\textbar} & \multirow{2}{*}{\textbar $E$\textbar} & \multirow{2}{*}{BKS}     & \multicolumn{3}{c|}{Random} & \multicolumn{3}{c|}{MCTS+Random} & \multicolumn{3}{c|}{Greedy-Random} & \multicolumn{3}{c|}{MCTS+Greedy-Random} & \multicolumn{3}{c|}{Greedy} & \multicolumn{3}{c}{MCTS+Greedy}                                                                                                          \\
                                                   &                                       &                                       &                          & best                        & avg                              & time                               & best                                    & avg                         & time                            & best & avg   & time & best        & avg   & time   & best & avg & time & best         & avg   & time   \\ \hline
                myciel6gb                          & 95                                    & 755                                   & 94                       & 190                         & 240.2                            & 0                                  & 100                                     & 110.1                       & 1                               & 104  & 116.5 & 0    & 96          & 97.9  & 2      & 99   & 99  & 0    & 98           & 98    & 0      \\
                myciel6g                           & 95                                    & 755                                   & 26                       & 52                          & 65.3                             & 0                                  & \textbf{26}                             & 28.9                        & 1                               & 28   & 31.1  & 0    & \textbf{26} & 26    & 0      & 27   & 27  & 0    & 27           & 27    & 0      \\
\end{tabular}
        }
    }
\end{table}

\subsection{Detailed results of MCTS coupled with local search procedures}\label{app:ls}

Tables \ref{tab:app7}- \ref{tab:app12} report the detailed results for the MCTS variants combined with the local search procedures (cf. Section \ref{sec:MCTS_greedy}).

\begin{table}[!hp]
    \scriptsize
    \caption{Results on pxx instances for LS and MCTS+LS methods.}
    \label{tab:app7}
    \Rotatebox{90}{
        \resizebox{1.7\textwidth}{!}{
            \begin{tabular}{cccc|ccc|ccc|ccc|ccc|ccc|ccc|ccc|ccc}
                \multirow{2}{*}{instance}         & \multirow{2}{*}{\textbar $V$\textbar} & \multirow{2}{*}{\textbar $E$\textbar} & \multirow{2}{*}{BKS}      & \multicolumn{3}{c|}{AFISA} & \multicolumn{3}{c|}{MCTS+AFISA} & \multicolumn{3}{c|}{TabuWeight} & \multicolumn{3}{c|}{MCTS+TabuWeight} & \multicolumn{3}{c|}{RedLS} & \multicolumn{3}{c|}{MCTS+RedLS} & \multicolumn{3}{c|}{ILSTS} & \multicolumn{3}{c}{MCTS+ILSTS}                                                                                                                                                                     \\
                                                  &                                       &                                       &                           & best                       & avg                             & time                            & best                                 & avg                        & time                            & best                       & avg                            & time   & best          & avg  & time   & best          & avg    & time & best          & avg  & time  & best          & avg  & time & best          & avg  & time \\ \hline
                p06                               & 16                                    & 38                                    & 565*                      & \textbf{565}               & 565                             & 0                               & \textbf{565}                         & 565                        & 0                               & \textbf{565}               & 565                            & 0      & \textbf{565}  & 565  & 0      & \textbf{565}  & 565    & 0    & \textbf{565}  & 565  & 0     & \textbf{565}  & 565  & 0    & \textbf{565}  & 565  & 0    \\
\end{tabular}
        }
    }
\end{table}

\begin{table}[!hp]
    \scriptsize
    \caption{Results on rxx instances for LS and MCTS+LS methods.}
    \label{tab:app8}
    \Rotatebox{90}{
        \resizebox{1.7\textwidth}{!}{
            \begin{tabular}{cccc|ccc|ccc|ccc|ccc|ccc|ccc|ccc|ccc}
                \multirow{2}{*}{instance}         & \multirow{2}{*}{\textbar $V$\textbar} & \multirow{2}{*}{\textbar $E$\textbar} & \multirow{2}{*}{BKS}     & \multicolumn{3}{c|}{AFISA} & \multicolumn{3}{c|}{MCTS+AFISA} & \multicolumn{3}{c|}{TabuWeight} & \multicolumn{3}{c|}{MCTS+TabuWeight} & \multicolumn{3}{c|}{RedLS} & \multicolumn{3}{c|}{MCTS+RedLS} & \multicolumn{3}{c|}{ILSTS} & \multicolumn{3}{c}{MCTS+ILSTS}                                                                                                                                                                               \\
                                                  &                                       &                                       &                          & best                       & avg                             & time                            & best                                 & avg                        & time                            & best                       & avg                            & time & best          & avg    & time   & best          & avg    & time & best          & avg    & time   & best          & avg    & time   & best          & avg    & time  \\ \hline
                r01                               & 144                                   & 1280                                  & 6724*                    & 6727                       & 6772.1                          & 425                             & \textbf{6724}                        & 6724                       & 2087.8                          & 6798                       & 6852.1                         & 0    & \textbf{6724} & 6724   & 1661.7 & 6726          & 6808.2 & 0    & \textbf{6724} & 6724   & 262.9  & \textbf{6724} & 6724   & 15.1   & \textbf{6724} & 6724   & 14.2  \\
\end{tabular}
        }
    }
\end{table}

\begin{table}[!hp]
    \scriptsize
    \caption{Results on DIMACS instances with proven optima for LS and MCTS+LS methods (1/2).}
    \label{tab:app9}
    \Rotatebox{90}{

        \resizebox{1.7\textwidth}{!}{
            \begin{tabular}{cccc|ccc|ccc|ccc|ccc|ccc|ccc|ccc|ccc}
                \multirow{2}{*}{instance} & \multirow{2}{*}{\textbar $V$\textbar} & \multirow{2}{*}{\textbar $E$\textbar} & \multirow{2}{*}{BKS} & \multicolumn{3}{c|}{AFISA} & \multicolumn{3}{c|}{MCTS+AFISA} & \multicolumn{3}{c|}{TabuWeight} & \multicolumn{3}{c|}{MCTS+TabuWeight} & \multicolumn{3}{c|}{RedLS} & \multicolumn{3}{c|}{MCTS+RedLS} & \multicolumn{3}{c|}{ILSTS} & \multicolumn{3}{c}{MCTS+ILSTS}                                                                                                                                                                     \\
                                          &                                       &                                       &                      & best                       & avg                             & time                            & best                                 & avg                        & time                            & best                       & avg                            & time   & best         & avg   & time   & best         & avg   & time  & best         & avg & time   & best         & avg   & time  & best         & avg   & time  \\ \hline
                DSJC125.9g                & 125                                   & 6961                                  & 169*                 & 170                        & 174.6                           & 21                              & \textbf{169}                         & 169.3                      & 968.6                           & \textbf{169}               & 170.5                          & 9      & \textbf{169} & 169.8 & 1445.4 & \textbf{169} & 169   & 0.7   & \textbf{169} & 169 & 0.6    & \textbf{169} & 169   & 1016  & \textbf{169} & 169   & 502.1 \\
\end{tabular}
        }}
\end{table}

\begin{table}[!hp]
    \scriptsize
    \caption{Results on DIMACS instances with known optima for LS and MCTS+LS methods (2/2).}
    \label{tab:app10}
    \Rotatebox{90}{
        \resizebox{1.7\textwidth}{!}{
            \begin{tabular}{cccc|ccc|ccc|ccc|ccc|ccc|ccc|ccc|ccc}
                \multirow{2}{*}{instance}         & \multirow{2}{*}{\textbar $V$\textbar} & \multirow{2}{*}{\textbar $E$\textbar} & \multirow{2}{*}{BKS}      & \multicolumn{3}{c|}{AFISA} & \multicolumn{3}{c|}{MCTS+AFISA} & \multicolumn{3}{c|}{TabuWeight} & \multicolumn{3}{c|}{MCTS+TabuWeight} & \multicolumn{3}{c|}{RedLS} & \multicolumn{3}{c|}{MCTS+RedLS} & \multicolumn{3}{c|}{ILSTS} & \multicolumn{3}{c}{MCTS+ILSTS}                                                                                                                                                              \\
                                                  &                                       &                                       &                           & best                       & avg                             & time                            & best                                 & avg                        & time                            & best                       & avg                            & time   & best         & avg   & time   & best         & avg   & time  & best         & avg & time & best         & avg & time  & best         & avg & time \\ \hline
                R100\_1g                          & 99                                    & 506                                   & 21*                       & \textbf{21}                & 22.1                            & 2082.5                          & \textbf{21}                          & 21.2                       & 1889.8                          & 23                         & 23                             & 0.5    & \textbf{21}  & 21    & 986.4  & \textbf{21}  & 21.9  & 25.5  & \textbf{21}  & 21  & 55.1 & \textbf{21}  & 21  & 73.2  & \textbf{21}  & 21  & 40.9 \\
\end{tabular}
        }
    }
\end{table}

\begin{table}[!hp]
    \caption{Results on DIMACS instances with unknown optima for LS and MCTS+LS methods (1/2).}
    \label{tab:app11}
    \Rotatebox{90}{
        \resizebox{1.7\textwidth}{!}{
            \begin{tabular}{cccc|ccc|ccc|ccc|ccc|ccc|ccc|ccc|ccc}
                \multirow{2}{*}{instance} & \multirow{2}{*}{\textbar $V$\textbar} & \multirow{2}{*}{\textbar $E$\textbar} & \multirow{2}{*}{BKS} & \multicolumn{3}{c|}{AFISA} & \multicolumn{3}{c|}{MCTS+AFISA} & \multicolumn{3}{c|}{TabuWeight} & \multicolumn{3}{c|}{MCTS+TabuWeight} & \multicolumn{3}{c|}{RedLS} & \multicolumn{3}{c|}{MCTS+RedLS} & \multicolumn{3}{c|}{ILSTS} & \multicolumn{3}{c}{MCTS+ILSTS}                                                                                                                                                                                \\
                                          &                                       &                                       &                      & best                       & avg                             & time                            & best                                 & avg                        & time                            & best                       & avg                            & time   & best         & avg    & time   & best          & avg    & time  & best          & avg    & time   & best         & avg    & time   & best         & avg    & time   \\ \hline
                C2000.5                   & 2000                                  & 999836                                & \textbf{2151}        & 2382                       & 2408.2                          & 3601                            & 2428                                 & 2428.6                     & 1389.4                          & 2348                       & 2355.8                         & 3327   & 2370         & 2376   & 1763   & 2166          & 2191.8 & 3364  & 2389          & 2404.9 & 1968   & 2438         & 2446.3 & 3499   & 2601         & 2607.5 & 1947.5 \\
\end{tabular}
        }
    }
\end{table}

\begin{table}[!hp]
    \caption{Results on DIMACS instances with unknown optima for LS and MCTS+LS methods (2/2).}
    \label{tab:app12}
    \Rotatebox{90}{
        \resizebox{1.7\textwidth}{!}{
            \begin{tabular}{cccc|ccc|ccc|ccc|ccc|ccc|ccc|ccc|ccc}
                \multirow{2}{*}{instance}         & \multirow{2}{*}{\textbar $V$\textbar} & \multirow{2}{*}{\textbar $E$\textbar} & \multirow{2}{*}{BKS}      & \multicolumn{3}{c|}{AFISA} & \multicolumn{3}{c|}{MCTS+AFISA} & \multicolumn{3}{c|}{TabuWeight} & \multicolumn{3}{c|}{MCTS+TabuWeight} & \multicolumn{3}{c|}{RedLS} & \multicolumn{3}{c|}{MCTS+RedLS} & \multicolumn{3}{c|}{ILSTS} & \multicolumn{3}{c}{MCTS+ILSTS}                                                                                                                                                                           \\
                                                  &                                       &                                       &                           & best                       & avg                             & time                            & best                                 & avg                        & time                            & best                       & avg                            & time   & best         & avg   & time   & best         & avg   & time   & best          & avg   & time   & best         & avg   & time   & best         & avg   & time   \\ \hline
                myciel6gb                         & 95                                    & 755                                   & \textbf{94}               & \textbf{94}                & 94                              & 69.2                            & \textbf{94}                          & 94                         & 104.2                           & 99                         & 99                             & 0      & \textbf{94}  & 94.1  & 947.5  & \textbf{94}  & 95.2  & 0      & \textbf{94}   & 94    & 10.1   & \textbf{94}  & 94    & 11.2   & \textbf{94}  & 94    & 11.2   \\
\end{tabular}
        }
    }
\end{table}

\end{document}